\documentclass[sn-mathphys,Numbered,iicol]{sn-jnl}% Math and Physical Sciences Reference Style
\usepackage{graphicx}%
\usepackage{multirow}%
\usepackage{amsmath,amssymb,amsfonts}%
\usepackage{amsthm}%
\usepackage{mathrsfs}%
\usepackage[title]{appendix}%
\usepackage{xcolor}%
\usepackage{textcomp}%
\usepackage{manyfoot}%
\usepackage{booktabs}%
\usepackage{algorithm}%
\usepackage{algorithmicx}%
\usepackage{algpseudocode}%
\usepackage{listings}%
\usepackage{caption}%
\usepackage{subcaption}%
\usepackage{tikz}%
\usepackage{acronym}
\usepackage{float,array}
\usepackage{booktabs}
\newcommand{\addnote}[3][]{#3}

\newcommand{\mcrot}[4]{\multicolumn{#1}{#2}{\rlap{\rotatebox{#3}{#4}~}}} 

\newcommand*{\twoelementtable}[3][l]%
{%  
    \begin{tabular}[t]{@{}#1@{}}%
        #2\tabularnewline
        #3%
    \end{tabular}%
}

\acrodef{STFT}{short-time Fourier transform}
\acrodef{iSTFT}{inverse short-time Fourier transform}
\acrodef{TF}{time-frequency}
\acrodef{DOA}{direction-of-arrival}
\acrodef{VAD}{voice activity detector}
\acrodef{ASR}{automatic speech recognition}
\acrodef{BIU}{Bar-Ilan University}
\acrodef{TCN}{temporal convolutional network}
\acrodef{SDR}{signal-to-distortion ratio}
\acrodef{SOTA}{state-of-the-art}
\acrodef{RIR}{room impulse response}
\acrodef{SNR}{signal-to-noise ratio}
\acrodef{DNN}{deep neural network}
\acrodef{DOA}{direction of arrival}
\acrodef{CNN}{convolutional neural network}
\acrodef{TF}{time frequency}
\acrodef{PReLU}{pre-exponential linear unit}
\acrodef{PIT}{permutation invariant training}
\acrodef{Sep-TFAnet}{Separation TF Attention
Network}
\acrodef{WER}{word error rate}
\acrodef{SER}{semantic error rate}
\acrodef{HWU}{Harriot Watt University}
\acrodef{SI-SDR}{scale-invariant signal-to-distortion ratio}
\acrodef{SPP}{speech presence probability}
\acrodef{NLU}{natural language understanding}
\acrodef{GCC-PHAT}{generalized cross-correlation phase-transform}
\acrodef{TDOA}{time-difference of arrival}

\acrodef{HRI}{human-robot interaction}
\acrodef{SER}{Speech emotion recognition}

\acrodef{RNN}{recurrent neural networks}
\acrodef{BLSTM}{bidirectional long short-term memory}
\acrodef{LSTM}{long short-term memory}
\acrodef{MFCC}{mel frequency cepstral coefficients}
\acrodef{RMS}{root mean square}
\acrodef{ZCR}{zero-crossing rate}
\acrodef{CLDNN}{convolutional long- short-term deep neural network}
\acrodef{FC}{fully-connected}
\acrodef{FFT}{fast Fourier transform}
\acrodef{STFT}{short-time Fourier transform}
\acrodef{t-SNE}{t-distributed stochastic neighbor embedding}
\acrodef{RAVDESS}{Ryerson audio-visual dataset of emotional speech and song}
\acrodef{IEMOCAP}{interactive emotional dyadic motion capture}
\acrodef{WA}{weighted accuracy}

\begin{document}

% \title{The SPRING Project: Deploying a Conversational Robot in a Hospital}
% \title{Multi-modal and multi-party social robots in a gerontological day-care hospital: achievements of the H2020 SPRING project}
\title{Socially Pertinent Robots in Gerontological Healthcare}


\author*[1]{\fnm{Xavier} \sur{Alameda-Pineda}}\email{xavier.alameda-pineda@inria.fr}
\equalcont{These authors contributed equally to this paper.}

\author[5]{\fnm{Angus} \sur{Addlesee}}\email{a.addlesee@hw.ac.uk}
\equalcont{These authors contributed equally to this paper.}

\author[5]{\fnm{Daniel} \sur{Hern\'andez Garc\'ia}}\email{d.hernandez\_garcia@hw.ac.uk}
\equalcont{These authors contributed equally to this paper.}

\author[1]{\fnm{Chris} \sur{Reinke}}\email{chris.reinke@inria.fr}
\equalcont{These authors contributed equally to this paper.}

\author[1]{\fnm{Soraya} \sur{Arias}}%\email{soraya.arias@inria.fr}
\author[4]{\fnm{Federica} \sur{Arrigoni}}%\email{federica.arrigoni@unitn.it}
\author[1]{\fnm{Alex} \sur{Auternaud}}%\email{alex.auternaud@inria.fr}
\author[8]{\fnm{Lauriane} \sur{Blavette}}%\email{lauriane.blavette@aphp.fr}
\author[4]{\fnm{Cigdem} \sur{Beyan}}%\email{cigdem.beyan@unitn.it}
\author[1]{\fnm{Luis} \sur{Gomez Camara}}%\email{email}
\author[3]{\fnm{Ohad} \sur{Cohen}}%\email{ohad.cohen@biu.ac.il}
\author[4]{\fnm{Alessandro} \sur{Conti}}%\email{alessandro.conti-1@unitn.it}
\author[8]{\fnm{Sébastien} \sur{Dacunha}}%\email{sebastien.dacunha@aphp.fr}
\author[5]{\fnm{Christian} \sur{Dondrup}}%\email{c.dondrup@hw.ac.uk}
\author[3]{\fnm{Yoav} \sur{Ellinson}}%\email{email}
\author[7]{\fnm{Francesco} \sur{Ferro}}%\email{francesco.ferro@pal-robotics.com}
\author[3]{\fnm{Sharon} \sur{Gannot}}%\email{email}
\author[6]{\fnm{Florian} \sur{Gras}}%\email{email}
\author[5]{\fnm{Nancie} \sur{Gunson}}%\email{email}
\author[1]{\fnm{Radu} \sur{Horaud}}%\email{email}
\author[4]{\fnm{Moreno} \sur{D'Incà}}%\email{moreno.dinca@unitn.it}
\author[6]{\fnm{Imad} \sur{Kimouche}}%\email{email}
\author[7]{\fnm{Séverin} \sur{Lemaignan}}%\email{severin.lemaignan@pal-robotics.com}
\author[5]{\fnm{Oliver} \sur{Lemon}}%\email{email}
\author[6]{\fnm{Cyril} \sur{Liotard}}%\email{email}
\author[7]{\fnm{Luca} \sur{Marchionni}}%\email{luca.marchionni@pal-robotics.com}
\author[3]{\fnm{Mordehay} \sur{Moradi}}%\email{email}
\author[2]{\fnm{Tomas} \sur{Pajdla}}%\email{email}
\author[8]{\fnm{Maribel} \sur{Pino}}%\email{email}
\author[2]{\fnm{Michal} \sur{Polic}}%\email{email}
\author[1]{\fnm{Matthieu} \sur{Py}}%\email{email}
\author[3]{\fnm{Ariel} \sur{Rado}}%\email{email}
\author[4]{\fnm{Bin} \sur{Ren}}%\email{email}
\author[4]{\fnm{Elisa} \sur{Ricci}}%\email{e.ricci@unitn.it}
\author[8]{\fnm{Anne-Sophie} \sur{Rigaud}}%\email{anne-sophie.rigaud@aphp.fr}
\author[4]{\fnm{Paolo} \sur{Rota}}%\email{paolo.rota@unitn.it}
\author[5]{\fnm{Marta} \sur{Romeo}}%\email{email}
\author[4]{\fnm{Nicu} \sur{Sebe}}%\email{niculae.sebe@unitn.it}
\author[5]{\fnm{Weronika} \sur{Siei\'nska}}%\email{email}
\author[3]{\fnm{Pinchas} \sur{Tandeitnik}}%\email{email}
\author[4]{\fnm{Francesco} \sur{Tonini}}%\email{francesco.tonini@unitn.it}
\author[1]{\fnm{Nicolas} \sur{Turro}}%\email{email}
\author[1]{\fnm{Timothée} \sur{Wintz}}%\email{email}
\author[5]{\fnm{Yanchao} \sur{Yu}}%\email{y.yu@hw.ac.uk}

\affil[1]{\orgdiv{RobotLearn Team}, \orgname{Inria at Univ. Grenoble Alpes}, CNRS, LJK, \orgaddress{655, \street{Avenue de l'Europe},  \postcode{38334}, \city{Montbonnot}, \country{France}}}

\affil[2]{\orgdiv{Czech Institute of Informatics, Robotics and Cybernetics}, \orgname{Czech Technical University in Prague}, \orgaddress{\street{Jugoslávských partyzánů 1580/3}, \postcode{160 00} \city{Dejvice},
\country{Czechia}}}

\affil[3]{\orgdiv{Acoustic Signal Processing Laboratory}, \orgname{Bar-Ilan University}, \orgaddress{ \city{Ramat-Gan}, \postcode{5290002}, \country{Israel}}}

\affil[4]{\orgdiv{Department of Information and Computer Science}, \orgname{University of Trento}, \orgaddress{\street{Via Sommarive 9}, \postcode{38123}, \state{Trento}, \country{Italy}}}

\affil[5]{\orgdiv{Interaction Lab, Mathematical and Computer Sciences}, \orgname{Heriot-Watt University}, \orgaddress{\city{Edinburgh}, \postcode{EH14 4AS}, \country{United Kingdom}}}

\affil[6]{\orgname{ERM Automatismes}, \orgaddress{\street{561 allée Bellecour},  \postcode{84200}, \city{Carpentras}, \country{France}}}

\affil[7]{\orgname{PAL Robotics}, \orgaddress{\street{C/ Pujades 77-79}, \postcode{08005}, \city{Barcelona}, \country{Spain}}}

\affil[8]{\orgdiv{Lusage Living Lab}, \orgname{Assistance Publique - Hopitaux de Paris}, \orgaddress{\street{54-56 Rue Pascal}, \postcode{75013}, \city{Paris}, \country{France}}}


\abstract{Despite the many recent achievements in developing and deploying social robotics, there are still many underexplored environments and applications for which systematic evaluation of such systems by end-users is necessary. While several robotic platforms have been used in gerontological healthcare, the question of whether or not an autonomous social interactive robot with multi-modal conversational capabilities will be useful and accepted in real-life facilities is yet to be answered. This paper is an attempt to partially answer this question, via two waves of experiments including patients and companions in a day-care gerontological facility in Paris with a full-sized humanoid robot endowed with social and conversational interaction capabilities. The software architecture, developed during the H2020 SPRING project, together with the experimental protocol, allowed us to evaluate the acceptability (AES) and usability (SUS) with more than 60 end-users. Overall, the users are receptive to this technology, especially when the robot's perception and action skills are robust to environmental clutter and flexible to handle a plethora of different interactions. We also report and discuss some concerns and general comments of the users.}

%%================================%%
%% Sample for structured abstract %%
%%================================%%


\maketitle

\section{Introduction}\label{sec:intro}

Social robots~\cite{leite2013social,breazeal2016social}\addnote[intro-socialrobot]{1}{, i.e.\ autonomous machines designed to interact and communicate with humans using social cues like voice, gestures, and facial expressions,} are not yet commonly found in our public spaces, despite this vision being an imminent reality over 25 years ago~\cite{thrun1998robots}. In addition to classic robotic skills, such as object avoidance during navigation, \emph{social} robots must be able to seamlessly communicate with multiple people through natural verbal and non-verbal interaction. Over the past decade, social robots have been tested in museums, airports, libraries, shopping malls, bars, and hospitals \cite{al2012furhat, keizer2014machine, robotics2018franny, foster2019mummer, vlachos2020robot, gunson2022developing,khaksar2023robotics}, reporting many positive findings. They have been used to successfully make sports and rehabilitation exercises more entertaining \cite{sackl2022social}, assist older adults in care facilities \cite{langedijk2020studying,stegner2023situated} and hospitals \cite{blavette2022robot,gonzalez2021social}, navigate and engage with people in public spaces (like concert halls, hotel lobbies, and shopping malls) \cite{langedijk2020studying,hahkio2020service,foster2019mummer}, and engage in multi-party interaction \cite{keizer2014machine,wagner2023comparing}. However, today's social robots are far from perfect as they are often run in a wizard-of-oz setup (with the researchers controlling the robot's navigation and dialogue manually) \cite{langedijk2020studying,stegner2023situated,wagner2023comparing}. Those that do function independently are stationary \cite{hahkio2020service,sackl2022social}, have limited dialogue capacity (rule-based or closed-domain) \cite{keizer2014machine,foster2019mummer}, or are designed for single user interactions that cannot be guaranteed in public spaces.
%, or %Additionally, public spaces are noisy, which causes speech recognition issues \cite{hahkio2020service}, and Social robots often 
% have to be augmented with better speakers, cameras, and microphones to enable even wizard-of-oz interactions \cite{stegner2023situated,wagner2023comparing,addlesee2023building}. 

% Unlike controlled laboratory settings, populated spaces 
Such spaces require more complex robotic skills and introduce new underexplored challenges~\cite{cooper2023challenges}. The robot must be able to fuse multi-sensory input to sense people and objects~\cite{xu2022transcenter}, tracking their positions as the robot navigates through the space~\cite{dubenova2022dinloc}, all while avoiding collisions. The robot must be able to hear its interlocutors~\cite{andriella2024dataset} despite background noise, the robot's ego-noise, the acoustics of the room, and multiple people speaking at the same time. It must understand where people are looking~\cite{tonini2023object}, and determine if they are getting frustrated with the robot to aid communication. The robot must move its head and eyes to look at its addressee or nod, move its arms appropriately when pointing, and move in the correct direction when guiding people. Typical speech systems are created to converse with only one individual at a time (e.g.\ Amazon Alexa, or Siri~\cite{berdasco2019user}), whereas pairs and groups of individuals may approach the robot together. People talk to each other as well as to the robot in a multi-party conversation, and the robot's spoken dialogue system must be able to handle this~\cite{addlesee2024multiparty}. Similarly, the navigation~\cite{pikuli2024navigating} or tracking~\cite{alameda2015vision} skills of the robotic platforms must be able to adapt to new environments without requiring extensive retraining or adaptation.

\addnote[intro-question]{1}{To tackle the above challenges, the EU's H2020 SPRING project\footnote{\url{https://spring-h2020.eu/}} aimed to develop social robots that can communicate in complex and unstructured public spaces. 
The SPRING (Socially Pertinent Robots IN Gerontological healthcare) project is a consortium of five international research labs (computer science \& engineering), two industry partners, and importantly, a gerontology hospital with research facilities. 
In this context, the main research question of the study presented in this paper is whether or not patients of a gerontology day-care hospital and their companions will find an autonomous social robot useful and acceptable in this environment.}

\begin{figure}[t]
    \centering
    \includegraphics[width=1.0\linewidth]{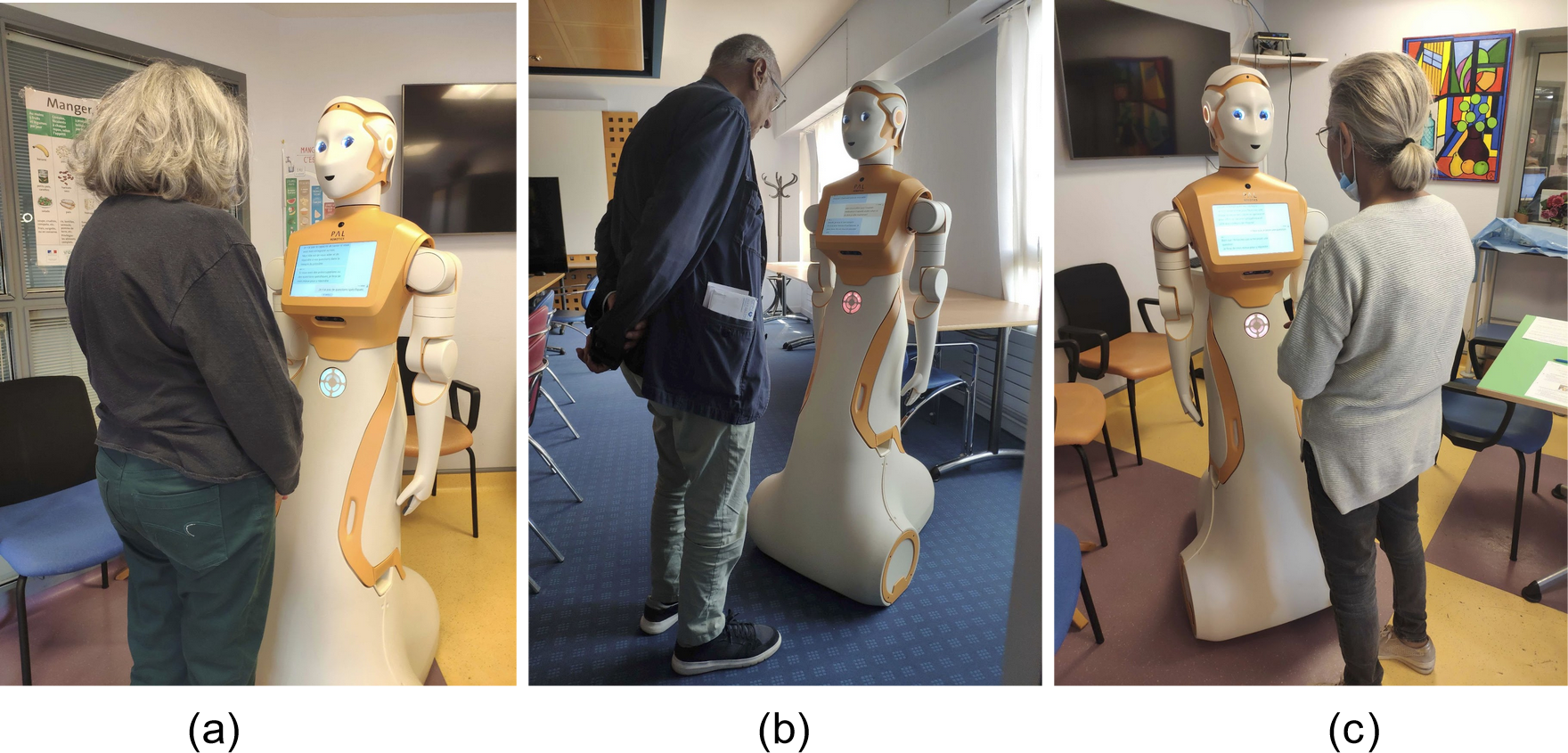}
    \caption{Examples of interactions between patients and the ARI robot in the Broca day-care hospital.}
    \label{fig:example-interaction}
\end{figure}

More precisely, our experimental setting is the Broca gerontology day-care hospital, where patients visit when they are suspected to have dementia. 
Patients typically spend full days at the hospital with a friend or family member for support. The hours are filled with multiple appointments, but a large portion of the day is also spent waiting for test results or the next appointment. 
In order to answer our research question, we have developed a robotic system able to autonomously provide participants with information and some light distraction from their otherwise stressful day, and we have conducted a long series of experiments with volunteer patients and their companions, interacting with the ARI robot, see Figure~\ref{fig:example-interaction}.

\addnote[intro-contr]{1}{The paper has three major contributions. First, a feature-rich robotic software architecture with advanced sensing and acting skills that enable a full-sized humanoid robot (ARI is 1.65~m tall) with autonomous social interaction capabilities. Second, an experimental protocol to evaluate this technology in a day-care gerontology hospital with real patients and companions, including ethical and practical considerations. Third, a series of experimental waves, involving more than 60 participants, conducted at the Broca hospital, their associated results, comparison and, discussion.} 
% Our study is unique in the sense that (i) the evaluation is conducted with real patients and companions of and in a day-care gerontological hospital (ii) with a full-sized humanoid robot (ARI is $1.65$~m tall) (iii) that is capable of advanced sensing and acting skills allowing it to converse with patients and companions. 
Our findings suggest that both patients and companions find such technology useful and acceptable, especially if it is robust to noise and clutter, and flexible to the plethora of different situations it can encounter.

The rest of the paper is structured into two main sections, namely Architecture and Experiments, before reaching the Conclusions. 
In Section~\ref{sec:architecture}, we detail the overall system architecture and provide technical details of each robotic `skill', and the improvement beyond the state-of-the-art. 
Some of the modules in the architecture were not evaluated in the hospital, but only in our laboratories due to Ethical constraints. 
Where appropriate, we refer to technical reports or publications to condense this paper without it becoming over-simplistic. 
\addnote[readers-guide]{1}{Readers mostly interested in the conducted experiments and the associated results are strongly encouraged to read the architecture overview in Section~\ref{sub:overall}, then jump to Section~\ref{sec:experiments}, where we describe the experimental setup in the hospital, including the ethical considerations, the protocol, the performance metrics, the main results, and the failure cases.} Finally, we conclude and list open topics in Section~\ref{sec:conc}.

\section{Architecture}\label{sec:architecture}

\subsection{Overview \& Robot Platform}\label{sub:overall}

\begin{figure*}[ht]
    \centering
    \includegraphics[width=\linewidth]{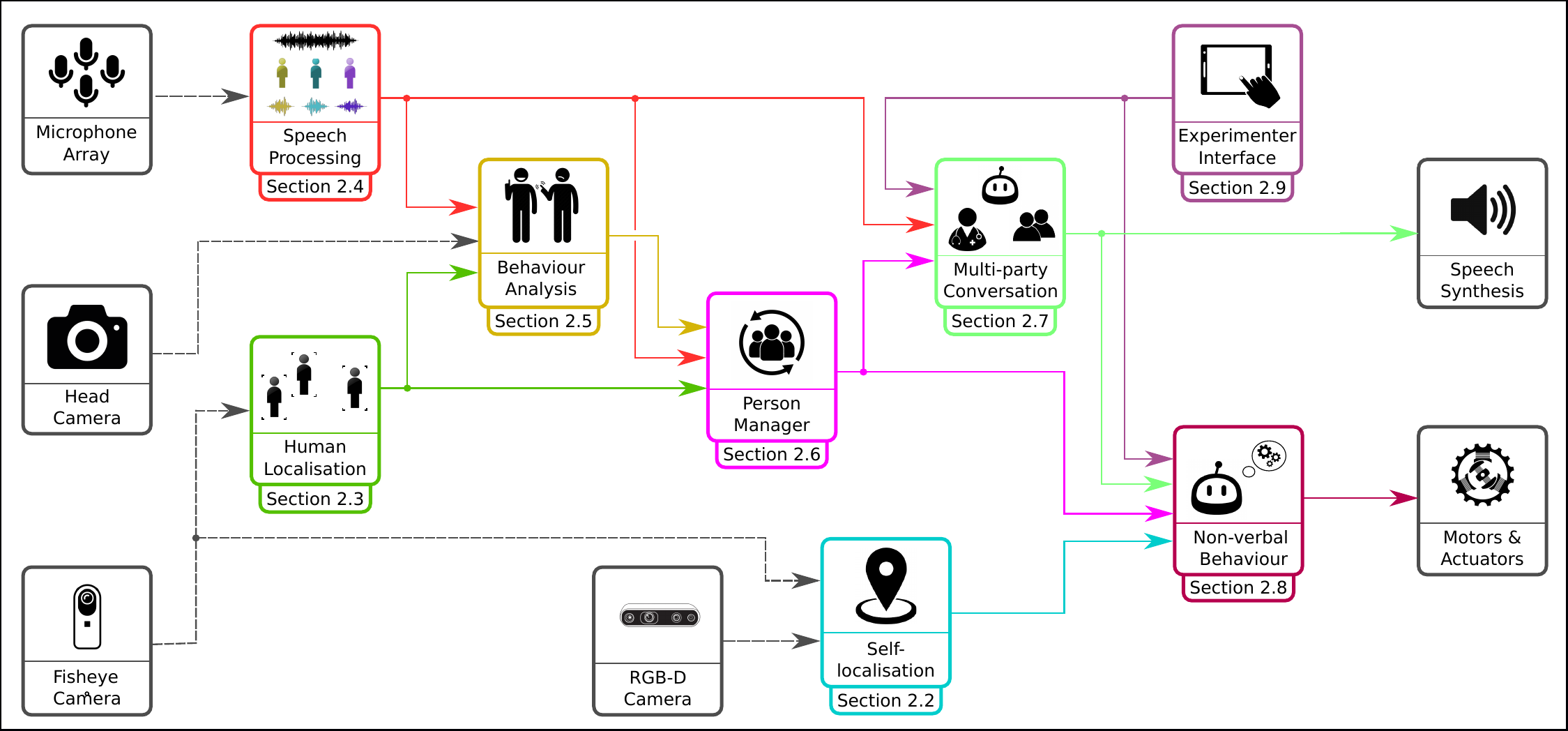}
    \caption{Overview of the SPRING architecture. Each box represents either a sensor (Microphone Array, Fisheye Camera, Frontal Camera, or RGB-D Camera), an actuator (Loudspeaker, Robot Motors), or a module (perception or action) corresponding to the subsections of Section~\ref{sec:architecture}.
    }
    \label{fig|arch}
\end{figure*}

To understand whether or not patients of a gerontology day-care hospital and their companions will find a social robot useful and acceptable, we developed and implemented a novel software architecture for humanoid robots. 
\addnote[novel-arch]{1}{Indeed, to our knowledge, the architecture proposed in this paper is the only one developed for enabling multi-modal social interactions with multiple persons in unscripted conversations.}
The architecture is composed of eight modules (Figure~\ref{fig|arch}) responsible for perception (self-localisation, human localisation, speech processing, human behaviour analysis, person manager) and action (experimenter interface, multi-party conversation, non-verbal behaviour generation) processes.
% Each module will be introduced in the following sections. 
They are developed on top of the ROS~1 Noetic\footnote{\url{http://wiki.ros.org/noetic}} middleware. 
In total, all modules consist of 52 ROS nodes communicating through more than 170 ROS topics and actions.  
Where relevant, the architecture uses standard ROS messages. 
%In particular, the SPRING architecture is one of the first to fully adopt the REP-155 `ROS4HRI'\footnote{\url{https://www.ros.org/reps/rep-0155.html}}~\cite{mohamed2021ros4hri,ros2023ros4hri} standard for human-robot interaction to combine the different perception modalities (voices, faces, bodies) into a consistent representation of persons (through the person manager, Section~\ref{sub:personManager}) that can be used by downstream action modules.
In particular, our architecture is one of the first to fully adopt the REP-155 `ROS4HRI'\footnote{\url{https://www.ros.org/reps/rep-0155.html}}~\cite{mohamed2021ros4hri} standard for human-robot interaction to combine the different perception modalities (voices, faces, bodies) into a consistent representation of the persons in the scene and %(through the person manager, Section~\ref{sub:personManager}) 
that can be used by downstream action modules.

The developed architecture is deployed on the PAL ARI humanoid robot, designed for use as a socially assistive companion~\cite{Cooper2020}. 
ARI is 1.65m tall, with a differential-drive mobile base, a touch-screen on the torso, movable arms to gesture, and a head with LCD eyes that enable expressive gazing behaviours. 
It is equipped with a four-microphone array (front torso), an RGB camera (head), and two 180$^{\circ}$ fish-eye cameras (chest and back) allowing us to capture and record the audio and video of the whole interaction from the robot's perspective. 
%It is equipped with a ReSpeaker Mic v2.0 array\footnote{\url{https://wiki.seeedstudio.com/ReSpeaker\_Mic\_Array\_v2.0}}, an RGB camera (in the head), and a 180$^{\circ}$ fish-eye camera (in the chest) allowing us to capture and record the audio and video of the whole interaction from the robot's perspective. 
An additional RGB-D camera is located in the front for self-localisation.   
The robot verbalises given responses using Acapela Text-To-Speech.\footnote{\url{https://www.acapela-group.com/}}
\addnote[our-arch]{2}{To our knowledge, the proposed architecture is unique in the sense that it follows several design requirements associated with our experimental goal: it exploits audio and visual data captured with the robotic sensors on board, it represents a social scene involving multiple people via the verbal and non-verbal behavioural cues, it is capable of deciding which social and dialogue actions are needed depending on the representation of the social scene, it has a usable experimenter interface, and finally all the computations are carried out on the premises. In addition, it is the first architecture effectively exploiting the ROS4HRI standard~\cite{mohamed2021ros4hri}.}

\addnote[general-arch]{1}{Generally speaking, our architecture can be used with other robotic platforms, provided that they have similar sensing and acting capabilities (audio-visual perception, self-localisation-oriented cameras, loudspeakers, and a mobile base). While, in practice, some minor adjustments will always be required, our modular structure allows for changes only in the few ROS modules that are in direct contact with the sensors and actuators, while keeping the rest untouched}. The following sections provide details of the key features of each module.

% Decision-making is performed using a combination of dialogue-driven behaviours, and plan-based behaviours, generated by a PetriNet. 
% \note{Daniel -> can you write this section, maybe?}

% Action execution, including autonomous navigation, is... \note{Chris/Xavi...}

\subsection{Self-Localisation}\label{sub:selfLocal}

The self-localisation module takes the RGB-D and Fisheye images as input and provides a 2D occupancy map indicating obstacles and the current position of ARI in the map.
This information is crucial for obstacle avoidance during the navigation of ARI (Section~\ref{sub:gestures}).
The developed module integrates RTABMap~\cite{labbe2019rtab}, a library for mapping and real-time tracking that incorporates ORB-SLAM~\cite{mur2017orb}, with an augmented version of the Hierarchical-Localisation (HLOC)~\cite{sarlin2019coarse} algorithm.
RTABMap and HLOC provide two distinct, pre-aligned maps that share the same global coordinate system. 
The pre-alignment is done by aligning the camera centres registered in these maps. 
The module fuses both information to provide a single consistent map for other modules, by aligning the camera centers in both maps.
This multi-layered approach is crucial for robust and precise localisation in complex and dynamic environments. %, as at Broca Hospital.

% The RTABMap framework,, initially provides efficient real-time tracking and mapping. 
RTABMap uses the RGB-D camera that is slightly down-faced at the front of ARI to provide in many situations efficient mapping and localisation.
However, certain environments might have reflective floors and RTABMap has trouble tracking key points, losing its localisation. 
For this purpose, an extended version of the HLOC algorithm was introduced to allow localisation based on the more robust and feature-rich images from the front and backwards-facing fisheye cameras. A calibration step, that allows to associated camera poses (position and orientation) and robot poses is done once-and-for-all at the beginning.   
Given a Fisheye image, the camera pose is initialised utilising the SENet~\cite{lee2023revisiting} approach for advanced image retrieval. 
Image retrieval estimates the camera's pose with an accuracy of a few meters, by identifying the most similar images from a collection of images of the environment with their poses in a global coordinate system.
The next step is to establish a set of correspondences between the input image and the closest images. 
%A critical subsequent step involves establishing 2D-3D correspondences between the ARI's fisheye camera images and the HLOC map, forming a comprehensive set of correspondences for camera rig localisation. 
The culmination of this process is the pose estimation using the generalised absolute pose solver~\cite{kukelova2016efficient} from PoseLib\footnote{\url{https://github.com/PoseLib/PoseLib}}, assuming known relative poses between the ARI cameras (i.e., camera rig) from external calibration. 

Combining state-of-the-art image retrieval and pose estimation techniques, this complex methodology results in state-of-the-art localisation accuracy. 
Using front and backwards-facing images captured from two consecutive time points (i.e., four fisheye images), the system achieves localisation within the maps of the Broca gerontology day-care hospital with less than 1cm positional error and 0.2 degrees rotational error in 90\% of queries. 
Comparatively, with a single image, the success rate drops to 15\% for positional accuracy within 1cm and 30\% for rotational accuracy within 0.2 degrees~\cite{spring-d24}. 
These findings highlight the enhanced precision of the multiview approach over single-image HLOC localisation used to initialise the real-time camera pose tracking performed by ORB-SLAM.

\subsection{Human Localisation}\label{sub:humanLocal}

The main goal of this module is to localise the humans in the space through time. To that aim, we use audio-visual data -- more precisely the front fisheye images and the microphone array signals -- to detect people, track them over time, and re-identify\footnote{Re-identification is a standard task in computer vision, aiming to understand whether two images are of the same or different persons, and has nothing to do with retrieving personal information of an individual from an image.} them (meaning that there is a one-to-one correspondence between persons and tracks). 
%the detection, re-identification, and tracking of people and groups over time using visual and auditory data. 
This information is exploited by the Conversation Manager (Section~\ref{sub:nlu}) to have a time-consistent identifier for each conversation partner and the human-aware navigation to move naturally around humans (Section~\ref{sub:gestures}).    
On the audio side, we employed a \ac{TDOA} estimation for the audio modality, utilising an instantaneous version of the \ac{GCC-PHAT}~\cite{knapp1976generalized}. 
The \ac{GCC-PHAT} algorithm reliably provides \ac{TDOA} readings in frames with a single speaker. 
This implementation utilises two horizontal microphones from the Respeaker microphone array embedded in ARI. 
The \ac{TDOA} readings are then translated into \ac{DOA} estimates using geometric considerations. 
On the visual side, we have implemented a state-of-the-art multi-person visual tracker known as fair multi-object tracking (FairMOT)~\cite{zhang2020fairmot}. 
This method combines the detection and the re-identification abilities and is based on the well-known residual neural network (ResNet34)~\cite{he2016deep}. 
However, the standard architectures are trained for regular cameras, while ARI's camera has a fisheye lens. This required an annotation and retraining procedure described in~\cite{spring-d32}. 
Another important property of FairMOT is that the tracking is based on the Kalman filter, plus an additional matching step that associates detections (from the current time step) and Kalman predictions (from the previous time step), by means of a detection-to-prediction distance matrix. By computing the distance between \ac{DOA} and the predictions of the Kalman filter, we can seamlessly incorporate audio detections to the tracking pipeline, thus achieving multi-modal speaker detection and tracking~\cite{alameda2011finding,ban2017exploiting}.
% This is important because as long as we can construct a distance matrix, we can associate not only visual detections but also audio ones. 

% , and these estimates are further enriched with spatial information obtained through visual means. 
% Additionally, depth information obtained through the visual modality proves valuable in accurately determining the \ac{DOA}, particularly in scenarios where speakers are close to the robot. 
%As briefly explained before, the TDOA estimators are associated with visual detections, 

Based on the position of humans in the image, we can extract their pose using OpenPose~\cite{cao2019openpose}. 
This provides the orientation of each person and their feet position in the image. 
By triangulation and the assumption that persons stand on an even floor, we can recover their depth. 
This allows to infer the distance and orientation to each other to finally detect conversational groups (group centre and its members) using the Graph-Cuts for F-formation (GCFF) algorithm \cite{setti2015f}.

%It is noteworthy that while the \ac{GCC-PHAT} algorithm reliably provides \ac{TDOA} readings in frames with a single speaker, a robust extension of the \ac{GCC-PHAT} \cite{Schwartz2023Array-Robust} will be explored for cases involving concurrent activity of multiple speakers, should an audio-based \ac{DOA} estimate be necessary for such scenarios.

\subsection{Speech Processing}\label{sub:speech}
% \note{Speech Processing [Sharon, Yoav, Pini, Ohad, Ariel, Mordechay BIU]}
% \note{Old Tex Version in old_tex/speech_processing.tex}

% \begin{figure*}[htbp]
%  \centering
% \input{biu/audio_pipeline}
%  \caption{Simplified Audio Pipeline of the Multi-party ASR}
%  \label{fig:pipeline}
% \end{figure*}
In real-life scenarios, a robot may engage with a group of people amid noisy and reverberant surroundings. The speech processing module's objective is to generate multiple streams of transcribed speech for all speakers from the microphone array signals, which the conversational system (Section~\ref{sub:nlu}) will utilise. 
The transcribed text streams should maintain consistency over time. This may require various methods of attributing identity to the speakers, including \ac{DOA} estimation and speaker identification.

We limit the scenarios to a maximum of two concurrent speakers. We also assume that the robot interacts with individuals in a half-duplex manner; namely, it does not listen while talking. The audio pipeline uses three steps to process the raw audio signal captured with the ReSpeaker microphone array:
\begin{enumerate}
% \item Acquire raw audio from the microphone array;
\item Enhance the speech quality;
\item Extract key attributes (voice activity, \ac{DOA}, speaker identity);
\item Transcribe the enhanced audio data.
\end{enumerate}

This process was designed to comprehensively handle the intricacies associated with single- and multi-speaker scenarios, using a systematic approach for audio processing and analysis.

% \paragraph{Raw audio acquisition}
% The audio is acquired via a ReSpeaker microphone array positioned in the belly of the ARI Robotic Platform. This choice was discussed internally, and several different options were considered. Several criteria such as audio quality, software/hardware interface, and physical setup possibilities within ARI were considered and described in a (confidential) deliverable~\cite{spring-d71}. The ReSpeaker microphone array outputs four raw microphone channels that we exploit as follows.

\paragraph{Speech enhancement and denoising}
The recorded speech signal is contaminated with various artefacts, such as noise and reverberation. To produce a clean speech signal, we apply the three following alternative speech enhancement algorithms. 

If speech is only contaminated by noise, a noise reduction module based on a Mixture of Deep Experts~\cite{Chazan2021MOE} will be activated. Each expert is implemented via a \ac{DNN} attuned to a distinct speech spectral pattern, such as a phoneme. Each expert generates a \ac{SPP} map, determining whether a time-frequency bin is predominantly speech or noise based on its expertise. The final time-frequency mask is derived by weighting the \ac{SPP} estimates from various experts and then applied to enhance the speech signal. 

If two speakers are active in the scene, a recent single-microphone speaker separation algorithm~\cite{Opochinsky23single} will be activated, aiming at noisy and reverberant signals typical of real-world environments. Two variants of the algorithm are available, \ac{Sep-TFAnet} and \ac{Sep-TFAnet}\textsuperscript{VAD}. 
Audio samples of the separation results (in English) are publicly available.\footnote{\url{https://Sep-TFAnet.github.io}} 

As an alternative, a speaker extraction module will be activated. We have implemented a two-stage method that extracts the speech corresponding to a reference signal and subsequently applies a dereverberation and residual interference suppression stage~\cite{Eisenberg2023}. A noteworthy feature of the speaker extraction algorithm, particularly pertinent to the SPRING project, is its capability to infer speaker embeddings that can be leveraged for speaker identification tasks. 

An arbitrator will be implemented to select the most appropriate algorithm based on the sound scene. Currently, the audio pipeline is tested and verified by manually switching between the alternative modules. 

\paragraph{Attribute extraction}
We extract three main speaker attributes: the activity pattern (if there is speech or not), the \ac{DOA}, and the identity (understood as something that characterises the voice rather than the name). While the first two attributes are short-term, identity is considered long-term.

Regarding the short-term attributes, the activity can be directly obtained from the Sep-TFAnet\textsuperscript{VAD} network, which incorporates an \ac{VAD}. The activity patterns of the separated speakers can serve for diarisation in the downstream dialogue manager. Short-term identification relies partly on spatial information derived from the current scene. This involves employing a late fusion mechanism that combines visual-based and audio-based \ac{DOA} estimation, as explained in Section~\ref{sub:humanLocal}. 

Regarding identity, we utilise Nvidia's ECAPA-TDNN model~\cite{dawalatabad2021ecapa} to extract voice-specific (speaker) embeddings, producing a 192-dimensional voice signature vector. %\footnote{\url{https://catalog.ngc.nvidia.com/orgs/nvidia/teams/nemo/models/ecapa_tdnn}} 
The speaker identification module then stores these embeddings in an internal database. The identification process occurs by comparing the cosine similarity between an active speaker and an entry in the database, triggering a match when the similarity exceeds a specified threshold. Finally, the speaker embedding obtained from the speaker extraction algorithm \cite{Eisenberg2023} may also serve as a speaker ID. %, at least during the conversation.

\paragraph{Audio transcription}
Transcribing audio is crucial in social robotics, as this component will generate the words comprehended by the robot. We carried out an extended experimental evaluation of several \ac{ASR} systems \addnote[asr-french]{1}{in French}~\cite{spring-d52} and concluded that the best option was to use Nvidia's on-prem solution called RIVA (Version 2.7). Apart from demonstrating performance comparable to existing cloud services, RIVA offers the advantage of adaptability to our data distribution. Moreover, it operates on-premises, mitigating potential concerns associated with data and privacy issues.

%%%%%%%%%%%%%%%%%%%%%%%%%%%%%%%%%%%%%%%%%%%%%%%%%%%%%%%%%%%%%%%%%%%%%%%%%%%%%%%%%%%%%%%%%%%%%%%%
%%%%%%%%%%%%%%%%%%%%%%%%%%%%%%%%%%%%%%%%%%%%%%%%%%%%%%%%%%%%%%%%%%%%%%%%%%%%%%%%%%%%%%%%%%%%%%%%
\subsection{Human Behaviour Analysis}\label{sub:behaviour}

This section describes the modules of ARI regarding the perception of human behaviour, namely: gaze target detection, detection of social acceptance of ARI, and multimodal emotion recognition. These modules input the position and groups of people from Section~\ref{sub:humanLocal}, the speech as processed in Section~\ref{sub:speech} as well as the face camera image, and could have direct applications in understanding the people's intentions, emotional state, and engagement with the robot. While we have not had the opportunity yet to exploit the outcome of these modules in hospital settings, they are part of the software architecture and publicly available as the other ones.

%Gaze target detection holds various applications in human-robot interaction, enabling robots to comprehend people's interests, intentions, or actions. Additionally, it can aid in identifying conversational groups by analysing who is gazing at whom. Emotion recognition, on the other hand, empowers robots to respond to their human partners, fostering more natural and effective interactions. Furthermore, robots equipped with emotion recognition can adjust their behaviour based on the user's emotional state. Our objective extends to automatically detecting the social acceptance of a robot by analysing the gaze behaviour of the human agent, providing the robot insights into whether the person is engaging with it. This capability can enhance the robot's adaptability, allowing it to propose improved dialogue, thereby attracting the human agent.

\paragraph{Gaze Target Detection} This task, also referred to as \emph{gaze-following} \cite{chong2020detecting,fang2021dual}, is to infer where each person in the scene (2D or 3D) is looking at \cite{chong2020detecting,fang2021dual,Recasens2015}. We aim to predict a person's gaze in an RGB scene image captured by the head camera of ARI. To do so, we apply our method \cite{tonini2022multimodal}, whose inputs are: (a) an RGB scene image, which contains the field of the view of the head camera of ARI; (b) an RGB face image, which is cropped from the RGB scene image, corresponding to the person whose gaze will be estimated and is extracted using a Multi-task Cascade CNN \cite{zhang2016joint}, and (c) a scene depth image obtained from monocular depth estimation network of \cite{ranftl2020towards}. 
A \textit{fusion and prediction module} concatenates scene, depth, and head features to obtain the two outputs of the proposed method, namely i) the gaze heatmap that is a 1-channel 2D matrix whose peak value represents the gaze coordinates and ii) the probability of the gaze target being inside or outside the scene.

\paragraph{Automatic Social Acceptance}
We address the task of automatically detecting the social acceptance of ARI as an engagement detection problem. Human-robot engagement detection refers to the process of identifying and understanding the level of interaction, involvement, or connection between humans and robots in a given context~\cite{Salam2015}. This involves analysing various cues, such as verbal and non-verbal communication, gestures, facial expressions, and other social signals, to determine the extent to which a person is actively engaged with or responsive to a robot~\cite{anzalone2015evaluating}. 
To address this task, our proposed method concentrates on analysing the gaze behaviour of human agents. 
We leverage the former discussed gaze target detection module to extract handcrafted features, drawing inspiration from \cite{beyan2017prediction}, which has demonstrated promising results in analysing multi-party conversations. 
Given a video clip lasting 10 seconds and the information of the gaze location in the scene, we extract the ratio of frames when the person is looking at ARI, gazing out of the field, or in regions visible by ARI's head camera.
These features are used to train a Deep Multilayer Perception along with the engagement annotation for the corresponding video clip.

% Given a video clip lasting 10 seconds and the information of the gaze location in the scene detected by our module, we define a gaze vector, which is a
% discretization of gaze location at each video frame. The instances of that vector can be: i) gazing at ARI, ii) gazing out of the field of the view of the ARI head camera, or iii) another person.
% Using this gaze vector, we extract the ratio of frames when the person is looking at ARI, gazing out of the field, or in regions visible by ARI's head camera. These features are used to train an MLP composed of 5 densely connected layers, with hidden dimensions set to $32$ and ReLU activation between each layer, along with the engagement annotation for the corresponding video clip.

\paragraph{Multimodal Emotion Recognition}

Facial expressions, in addition to a person's speech (e.g. prosody, pitch), are an essential part of non-verbal communication and major indicators of human emotions. Effective emotion recognition systems can facilitate comprehension of an individual's intention, and prospective behaviours in Human-Robot Interaction. Our approach incorporates both a facial emotion recognition (FER) system~\cite{d2023unleashing}, designed to differentiate between \emph{positive} and \emph{negative} emotions, and a single-microphone speech emotion recognition (SER) system~\cite{Sherman2022} that can identify discrete emotions (e.g. happy, angry, sad).

\begin{figure*}[t!]
    \centering
    \includegraphics[width=\linewidth]{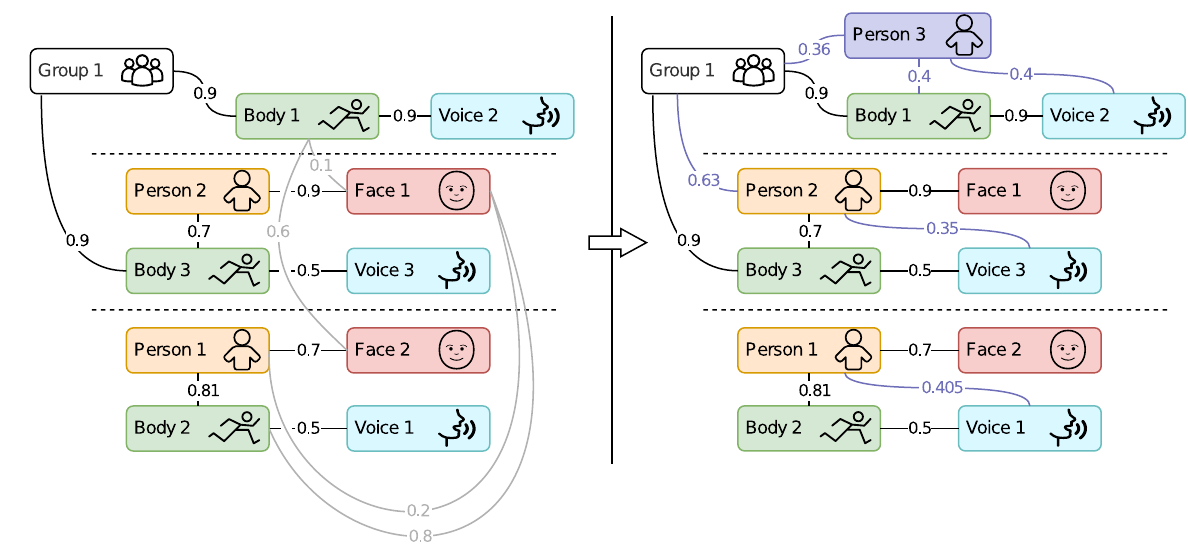}
    \caption{Example of the processing of social features by the person manager. Faces, voices, bodies, persons, and groups are denoted by different colours. Edge numbers represent the likelihood of two nodes being associated. The relations graph generated by the perception module is shown on the left, and the changed graph by the person manager to maximise the overall likelihood on the right. Gray edges on the left are discarded and purple edges and nodes on the right are added by the person manager. For instance, \texttt{Person 3} was added for features \texttt{Body 1} and \texttt{Voice 2} to be associated with a person. Additional edges were added to ensure stability over time (e.g. robustness against undetected bodies, faces, or voices).
     % (left) 
     % Illustrative example of a relations graph generated by the perception modules. 
     % Numerical values are likelihoods that two features belong to the same person. 
     % (right) 
     % The person manager partitions the graph to maximise overall likelihood.
     % One anonymous person had to be created for the features \texttt{Body~1} and \texttt{Voice~2}, as they would otherwise be dangling. 
     % Note that the \texttt{Person 1} - \texttt{Voice 1} edge, for example, does not exist in the original data.
     % It is a \emph{computed edge}, used to ensure stability if \texttt{Body 2} is not detected any more and disappears.
    }
    \label{fig:person_manager}
\end{figure*}

The FER system is composed of two neural networks: 1) a convolutional autoencoder-based architecture used as a feature extractor, and 2) the classification head responsible for classifying whether the emotion is positive or negative. The employed convolutional autoencoder is trained to reconstruct the input image (unsupervised pre-training). We freeze the autoencoder and use it only to extract features that are passed to a multilayer perceptron used as a linear classifier. This classifier is trained with focal loss to better handle class imbalance problems, if any, and classifies whether a given face is displaying a positive or negative emotion~\cite{d2023unleashing}.

Our SER algorithm~\cite{Sherman2022} is a variant of a previous single-microphone system~\cite{huang2017deep} to best fit the robot's hardware in our setting. The acoustic features are extracted from the audio utterances and fed to a neural network that consists of CNN layers, a BiLSTM combined with an attention mechanism layer \cite{bahdanau2014neural}, and a fully connected layer.

Both the FER and SER systems have been evaluated individually on relevant corpora (see~\cite{d2023unleashing} and~\cite{Sherman2022} respectively) achieving state-of-the-art performance, but there still exists a challenge in applying these models together in the real hospital. Indeed, the domain shift between the corpora and hospital data distributions is substantial and collecting annotated data in hospital settings is challenging and resource-consuming.

\subsection{Person Manager}
\label{sub:personManager}

The perception modules presented in Sections~\ref{sub:humanLocal} (human localisation),~\ref{sub:speech} (speech processing) and~\ref{sub:behaviour} (human behaviour analysis) all extract what we call social features.
To be used for downstream tasks (like the multi-party conversation manager, as described in the following section), these features have to be combined and associated with each other to build complete \emph{persons}.
The association might be direct (e.g., the facial recognition software module directly associates a face to a specific person), or indirect (we associate a body to a face based on the overlapping regions of interest in the source image, and transitively associate the body to a person).
To broadcast possible associations between features and/or persons, with their corresponding likelihoods, the SPRING architecture uses the ROS4HRI~\cite{mohamed2021ros4hri} standard.

ROS4HRI defines five types of entities to model the humans interacting in the vicinity of the robot.
The first three are feature entities for the \emph{face}, \emph{body} and \emph{voice}.
The last two are for \emph{persons} and \emph{groups}.
Each entity has a unique identifier and properties, e.g.\ the bounding box of a face or the position of the group centre.
The identifiers of feature entities are transients and might be created or changed at any time, based on the result of the face, body and voice detection algorithms. 
The \emph{person} class, has instead a persistent identifier: a given person should always get assigned the same \emph{person} identifier. 
Recognition modules are in charge of associating feature identifiers to the corresponding person identifier.
For instance, the facial recognition node might broadcast a message like [\{\texttt{john}, \texttt{face\_432}, $0.8$\}, \{\texttt{jane}, \texttt{face\_432}, $0.2$\}] to indicate that a detected face has 80\% chance of being John, and 20\% chance of being Jane.
This design allows the effective separation of concerns, where the question of `feature' (face, body, voice) detection can be cleanly isolated from the question of feature matching.
For SPRING, three feature matching processes are used: body to face, body to voice, and body to group.
Associations depend in general on the closeness of two features, for instance, how far a detected face is from a body, and the Hungarian algorithm~\cite{kuhn1955hungarian} is used to solve the assignment problem if several features have to be matched.
Please note, due to ethical considerations during the experiments in the hospital, a direct association of a person entity to an actual person is not done (e.g.\ through a facial mapping with a database of photos).
All person entities were regarded as anonymous persons.

This association mechanism is exploited to span, over time, a probabilistic graph of relationships between social features, persons and groups (Figure~\ref{fig:person_manager}). 
The challenge is then to compute the most likely person--features associations. %, also accounting for the fact that the person `owning' the feature might yet be unknown (for instance, a module detects a face, but the person is not yet recognised).
We have developed a novel algorithm, playfully named \emph{Mr. Potato algorithm}~\cite{lemaignan2024mrpotato}, to compute the most probable associations between the different features and persons.
Our algorithm searches all possible partitions of the graph, and selects the one that minimises the number of associations, while maximising \emph{affinity}, i.e. the sum of likelihoods of each association.
% In practice, our algorithm yields the three associations presented in Figure~\ref{fig:groups} (assuming in this example a likelihood threshold of $0.4$).
Our implementation represents efficiently the persons-features graph by the boost graph library~\cite{siek2001boost}. %\footnote{https://www.boost.org/} 
%Connected components are computed using \texttt{boost}'s \texttt{connected\_components} algorithm that uses a Depth-First-Search approach; likewise, minimum spanning trees are calculated using the \texttt{boost} implementation of the Kruskal's algorithm; and shortest paths between nodes are computed using the \texttt{boost} implementation of the Dijkstra algorithm.
Connected components are computed using a Depth-First-Search approach; likewise, minimum spanning trees are calculated using the Kruskal's algorithm~\cite{kruskal1956shortest}; and shortest paths between nodes are computed using the Dijkstra algorithm~\cite{dijkstra2022note}.
The result of the algorithm is published as a new set of ROS4HRI-compatible topics, listing the list of tracked and known groups and persons, and their corresponding face, body and voice. % as appropriate.

% \begin{figure}
%      \centering
%      \begin{subfigure}[b]{\columnwidth}
%          \centering
%          \includegraphics[width=\textwidth]{g1.png}
%          \caption{Group 1}
%          \label{g1}
%      \end{subfigure}
%      \hfill
%      \begin{subfigure}[b]{0.7\columnwidth}
%          \centering
%          \includegraphics[width=\textwidth]{g2.png}
%          \caption{Group 2}
%          \label{g2}
%      \end{subfigure}
%      \begin{subfigure}[b]{0.9\columnwidth}
%          \centering
%          \includegraphics[width=\textwidth]{g3.png}
%          \caption{Group 3}
%          \label{g3}
%      \end{subfigure}

%     \caption{
%     % Partition of the graph in Figure~\ref{fig:example} that maximise overall likelihood. 
%     Partition of the graph in Figure~\ref{fig:person_manager} that maximise overall likelihood.
%     One anonymous person had to be created in Group 3 for the features \texttt{body1} and \texttt{voice2}, as they would otherwise be dangling. 
%     Note that the \texttt{person1}--\texttt{voice1} edge in Group 2 does not exist in the original data: it is a \emph{computed edge}, used to ensure stability if \texttt{body2} is not detected any more and disappears.
%     }
%     \label{fig:groups}
% \end{figure}

\begin{figure*}[t]
\centering\footnotesize
\begin{tabular}{cclc}
\toprule
\textbf{Example} & \textbf{User} & \textbf{Utterance}  &\textbf{Note of Interest} \\ \midrule
\multirow{2}{*}{(A)} & U1   & I think it is London  & \multirow{2}{*}{\begin{tabular}[c]{@{}c@{}}If turn 2 was U2, it would be agreement,\\ so speaker recognition changes meaning.\end{tabular}} \\
                     & U1   & Yeah... London & \\\midrule
(B)                  & U1   & My husband needs the bathroom & Providing other user's goal. \\ \midrule
\multirow{2}{*}{(C)} & U1   & What time is my appointment? & \multirow{2}{*}{\begin{tabular}[c]{@{}c@{}}U2 answers U1's question, but addressee\\ was ambiguous without gaze info.\end{tabular}} \\
                     & U2   & It's at 10am & \\ \midrule
\multirow{2}{*}{(D)} & U1   & We are hungry & \multirow{2}{*}{\begin{tabular}[c]{@{}c@{}}Shared goal indicated by `we', and robot\\ can point to the `left'. Fasting is in red as\\ it is a world-knowledge hallucination.\end{tabular}} \\
                     & ARI  & \begin{tabular}[c]{@{}l@{}}The café is through the door on your left,\\ \textcolor{red}{but you should fast before your visit}.\end{tabular} & \\ \midrule
\multirow{4}{*}{(E)} & U1   & Name a song by... & \multirow{4}{*}{\begin{tabular}[c]{@{}c@{}}This is an OOD question that could not\\ be answered without the LLM-based\\ SDS. The partial utterance is handled\\ naturally which improves accessibility.\end{tabular}} \\
                     & ARI  & By who? & \\
                     & U1   & Queen & \\
                     & ARI  & Bohemian Rhapsody & \\ \bottomrule
\end{tabular}\vspace{2mm}
\caption{Examples of multi-party conversations from MPCs with the ARI social robot \cite{addlesee2023multiparty,schauer2023detecting}.}
\label{fig:mpcExamples}
\end{figure*}

%%%%%%%%%%%%%%%%%%%%%%%%%%%%%%%%%%%%%%%%%%%%%%%%%%%%%%%%%%%%%%%%%%%%%%%%%%%%%%%%%%%%%%%%%%%%%%%%
%%%%%%%%%%%%%%%%%%%%%%%%%%%%%%%%%%%%%%%%%%%%%%%%%%%%%%%%%%%%%%%%%%%%%%%%%%%%%%%%%%%%%%%%%%%%%%%%
\subsection{Multi-party Conversation Manager}\label{sub:nlu}

Commercial and research spoken dialogue systems (SDSs), conversational agents, and social robots have been designed with a focus on dyadic interactions. That is a two-party conversation between one individual user and a single system/robot. Such interactions can only be guaranteed in specific settings, like people interacting with Siri on their phone, or with Amazon Alexa in single-occupant homes. When Alexa is in a family home, its lack of multi-party capabilities is apparent \cite{porcheron2018voice}, but this becomes a critical limitation when deploying social robots in public space~\cite{al2012furhat, keizer2014machine, robotics2018franny, foster2019mummer, vlachos2020robot, gunson2022developing}, where multi-party conversations (MPCs), involving people talking to both the robot and each other, do commonly occur.

Tasks that are typically trivial in the dyadic setting become considerably more complex when conversing with multiple users \cite{traum2004issues, gu2022says}: (1) The speaker is no longer simply the other person, so the meaning of the dialogue depends on recognising who said each utterance (see (A) in Figure~\ref{fig:mpcExamples}); (2) addressee recognition is similarly more complicated as people can address each other, the robot, and groups of individuals; and (3) response generation depends on who said what to whom, relying on the semantic content and surrounding multi-party context. To make things even more difficult, MPCs provide additional unique challenges that are underexplored. Dyadic SDSs must identify and answer the user's goals to be practically useful. In MPCs, users can provide another person's goal (see (B) in Figure~\ref{fig:mpcExamples}), answer each other's goals (see (C) in Figure~\ref{fig:mpcExamples}), and even share goals (see (D) in Figure~\ref{fig:mpcExamples}, \cite{eshghi2016collective}). In SPRING, we have established the task of multi-party goal-tracking \cite{addlesee2023multiparty}.

The conversational system \cite{addlesee2024a,addlesee2024multiparty} has been iteratively improved through regular user tests and interviews with patients visiting the Broca Gerontology day-care hospital. The initial system \cite{gunson2022developing} was developed before recent LLM advances (such as ChatGPT), relying on a `traditional' modular architecture based upon Alana V2 \cite{papaioannou2017alana,curry2018alana}. As patients were usually accompanied by a companion, the lack of multi-party capabilities proved problematic. The system interrupted users since it responded to every turn, not allowing them to talk to each other at any point. We therefore designed and ran a multi-party data collection in a wizard-of-oz setup \cite{addlesee2023data, addlesee2023multiparty} (see Section \ref{sec:experiments}) and have used this data to motivate and evaluate our current SDS. Not only is this new system multi-party and multimodal, it improves QA accuracy, improves accessibility for people with dementia \cite{addlesee2024multiparty}, and enables added functionality. For example, where previously we had to specifically design the system to tell jokes and run entertaining quizzes \cite{addlesee2023detecting, schauer2023detecting}, LLMs can now handle this inherently due to their general knowledge. Most importantly, both users and the hospital staff have reported that the user experience has improved dramatically (see Section~\ref{sec:experiments}). 
%A video of our multi-party system and SDS architecture is available online.\footnote{\url{https://www.youtube.com/watch?v=xMCpcsLhN_I}}

%%%%%%%%%%%%%%%%%%%%%%%%%%%%%%%%%%%%%%%%%%%%%%%%%%%%%%%%%%%%%%%%%%%%%%%%%%%%%%%%%
\subsection{Non-verbal Behaviour Generation}\label{sub:gestures}
% \note{Behaviour Generation [Chris, Daniel, INRIA \& HWU], Social navigation Social MPC [INRIA, Chris]}

% general composition: behaviour generator and manager

Gaze, gestures, and navigation of ARI are controlled by two modules:
The \textit{behaviour manager} interfaces a high-level planner with the conversational system to choose appropriate actions during an interaction.
The \textit{behaviour generation} module provides and executes the actions.

\paragraph{Behaviour Manager}

The behaviour manager handles the interface between the \textit{conversation manager} and the \textit{behaviour generator}, and is responsible for deciding appropriate high-level social actions and managing the interactions. 

To enable situated interactions, \addnote[situated-interactions]{2}{meaning interactions anchored in the physical and social environments of the robot}, with multiple users at the same time, it is required that the non-verbal behaviour system components interface the social perception signals (presented in Sections \ref{sub:humanLocal}, \ref{sub:speech}, \ref{sub:behaviour} and \ref{sub:personManager}) with the multi-party conversational manager (presented in Section \ref{sub:nlu}). 
Among the set of social decisions that are required to be made by the non-verbal behaviour system, we have: detect people's arrival and departure, determines a person in the scene that wants or requires the robot's attention, decide when to go, start an approach or guidance action, decide who to look at, switch the focus of attention during multi-party interactions, etc.

The behaviour manager is implemented as an abstract controller for the ROS Petri-Net Planner (PNP)~\cite{Dondrup2019}. A Petri-Net is a mathematical model for state machines. %It defines places, transitions, arcs, and execution tokens. 
The behaviour manager can start and stop tasks in Petri-Net plans, and keep track of them. The implementation of the PNP supports automatic generation of Petri-Net machines, handles concurrent execution of multiple Petri-Net machines, and natively exploits ROS infrastructure. The PNP is fully implemented in ROS. 
It consists of the Petri-Net plan server, the knowledge-base (KB), a set of ROS action servers from the Petri-Net plans (recipes). 
The PNP starts tasks and keeps track of them. It is able to manage the currently available and running Petri-Nets and to provide functionality to send and receive information from/to a specific net. The other major functionality of the controller interface is to exchange data between the different plans and the social state representation. 

The main functionality is to manage the currently available and running plans and to provide the functionality to send and receive information between different plans and the social state representations provided by the social scene understanding components.
Through the interface with the social perception components (defined in the previous sections), the behaviour manager populates and maintains the planner's knowledge base with information about the interaction, the social state, the people engaged in the interaction/conversation with the robot, etc. 
The behaviour manager module interfaces with the non-verbal behaviour generation through the ROS actions servers that control low-level action execution (gestures, navigation) for a pertinent social interaction.

% \note{TODO: HWU}

\paragraph{Behaviour Generator}

\begin{figure}[t]
\centering
\includegraphics[width=0.9\linewidth]{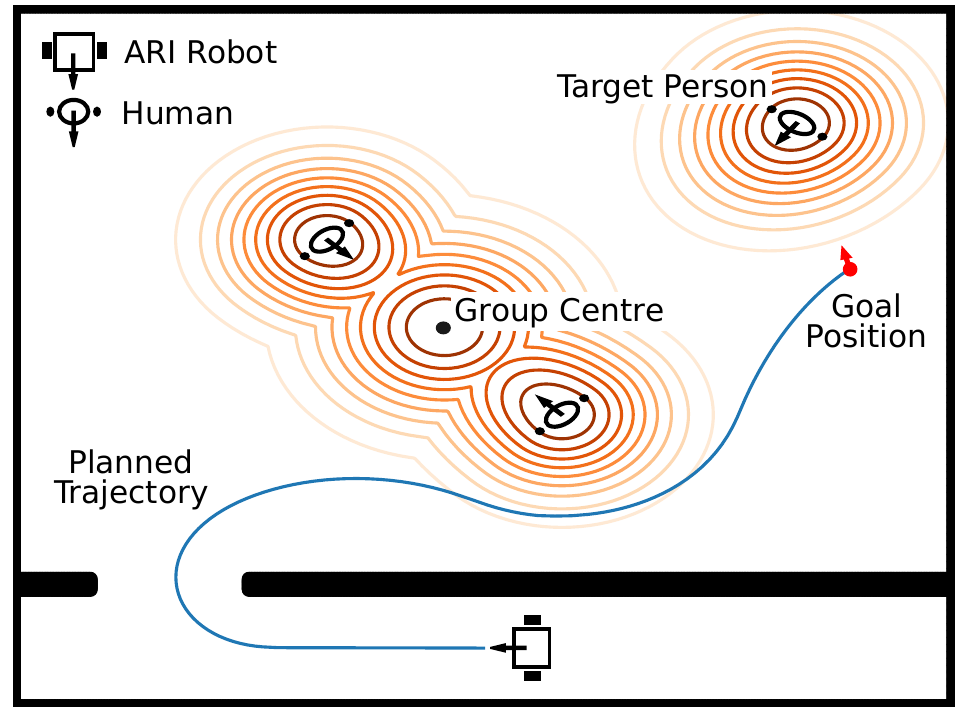}
\caption{The Social Navigation Controller integrates the social space of persons and groups (coloured contours indicate their cost) to plan a trajectory (blue) towards the target person without disturbing the conversational group.}
\label{fig:social_mpc}
\end{figure}

% behavior generator
The behaviour generator provides mainly actions and behaviours for two aspects:
First, it controls ARI's arms, head, and eyes to generate gestures such as waving or pointing and its gaze.
The gesture and gaze controllers are hard-coded behaviours that can be called during dialogues.

\begin{figure*}
    \centering    
    \includegraphics[width=0.8\textwidth]{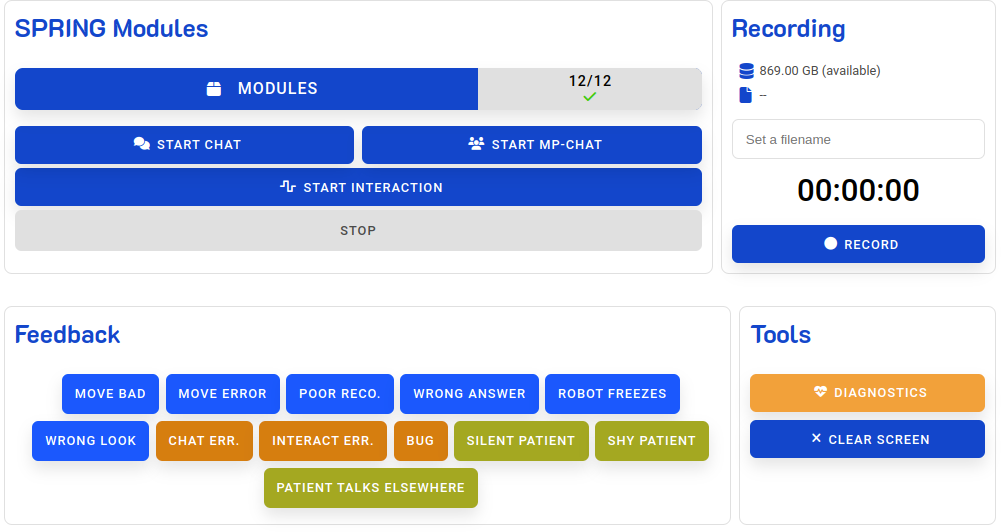}
    \caption{Snapshot of the Experimenter Interface.}
    \label{fig:control_screen}
\end{figure*}

Second, it provides a human-aware navigation controller to move among humans and join them for interactions. 
The controller faces two crucial challenges.
Foremost, it needs to move safely and reliably due to its application around vulnerable persons.
Secondly, it has to adhere to social norms regarding groups, for instance, respecting conversational groups and not moving through them, for example.
In difference to existing methods \cite{kruse2013human, mavrogiannis2023core, singamaneni2023survey}, we opted to combine a Model Predictive Controller~\cite{camacho2007model} with an explicit social space model of humans and groups \cite{truong2016dynamic}, therefore called Social Navigation Controller (SNC)~\cite{spring-d63}.
The SNC allows planning ARI's trajectory precisely over a future time horizon while adhering to constraints.
This provides a higher level of safety compared to other planning methods such as Dynamic-Window Approaches \cite{truong2017approach} or learned end-to-end controllers such as by Reinforcement Learning~\cite{chen2020robot}, see~\cite{pikuli2024navigating} for a broader perspective.  
Additionally, having an explicit social space model allows us to understand and tune it reliably compared to learned black-box models.

The SNC utilises a dynamic forward model $x_{t+1} = f(x_t, u_t)$ of ARI to predict its trajectory over a 2 sec time horizon, considering its current state $x_{1}$ (2D position in the map) and the anticipated motor outputs $u_\tau = \{u_{1}, \ldots, u_{T}\}$ (linear and angular velocity).
By optimising a loss function $J(x_1, u_\tau)$ that captures the desired performance and constraints, the SNC computes an optimal control trajectory for the motor outputs.
The first control action $u_{1}$ of this trajectory is applied and the optimisation process is repeated in the next time step, incorporating updated measurements and adjusting the control actions.
The loss function $J$ incorporates constraints about maximum velocities and a cost function. % that is composed of different components.
The costs are based on the occupancy map for obstacles (Section~\ref{sub:selfLocal}) and the position of humans and groups (Section~\ref{sub:humanLocal}).
Costs are high for areas close to objects and for interfering with the social space of humans and groups.
Social spaces are modelled around the position of humans and group centres by 2D Gaussian-like functions that are conditioned on their orientation, movement direction, and status (e.g. seated vs standing) following \cite{truong2016dynamic} (Figure~\ref{fig:social_mpc}).
The SNC avoids navigating through areas that incur a high cost, effectively avoiding obstacles, moving through groups, or being too close to humans.
Social spaces also define the position to join a person or group to start a dialogue following the approach in \cite{truong2017approach} by identifying the closest point to a group centre or a single person who is least interfering with their social space.

\subsection{Experimenter Interface}\label{sub:interface}
% \note{Experimenter Interface [ERM]}

During the experimentation, the robot will not be accompanied by computer science researchers or engineers, but by medical researchers and personnel. In order to conduct the experimental session, an appropriate experimenter interface was developed. Through a conception-trial-update process, we converged on an interface design that is a trade-off between usability and controllability. The interface (Figure~\ref{fig:control_screen}) is implemented for use on a dedicated tablet.

From a technical point of view, the interface connects to its server that runs on the external computer. This server communicates with the modules described above via ROS, interacts with the tablet via web technologies, and works as a gateway between these two parts. Additionally, it allows the experimenter to control the data collection. Via a web browser and the tactile screen, the experimenter can easily check the status of the robot and external computer, conduct the experiment, control the interaction as well as generate quick annotations of the interaction that are synchronised with the rest of the ROS messages. These annotations enable the project engineers to measure error rates, and diagnose bugs more easily. More details about the experimenter interface and other user interfaces (e.g.\ for data collection) can be found in~\cite{spring-d53}.

\section{Experiments}\label{sec:experiments}
This section presents the experiments to validate the acceptability and usability of the introduced multi-modal conversational system on the ARI robot.
The validation was conducted with real patients and companions at the Broca gerontology day-care hospital. 
The experimental protocol, measures, procedure, obtained results, as well as associated discussion and failure cases, are described in the following.

\subsection{Experimental Protocol}\label{sub:protocol}

\paragraph{Participants}

The study was carried out between May 2023 and January 2024. Two groups of participants were recruited for this study: elderly outpatients from a geriatric hospital and their accompanying persons. These participants were recruited from the Broca Gerontology day-care hospital in Paris.
Inclusion criteria for this study were: For patients (1) to be aged 60 and over, (2) have a good understanding of the French language, (3) not to have severe cognitive impairment (MMSE~$>$~10, see~\cite{folstein1975mini}) or neuropsychiatric symptoms (delirium, hallucination), and for accompanying persons: (1) be of legal age and (2) have a good understanding of the French language. All the participants were required to express their consent to participate in the study.

\paragraph{Ethical Approval and Data Availability Statement}

This research was fully supported by the H2020 SPRING Project funded by the European Commission. As such, there are no conflicts of interest to be disclosed. As per the research involving humans and the informed consent form, we provide details about the sought Ethical Committees and obtained Ethical Approvals in the following.

The study was approved by the French National Ethical Review Board \textit{Comité National de Protection des Personnes}, \textit{CPP Ouest II}, \textit{Maison de la Recherche Clinique-CHU Angers} (approval number: 2021/20), the local ethics committee of the University of Paris (\textit{Comité d'Ethique de la Recherche} CER-N IRB: 00012020-108), and was compliant with the General Data Protection Regulation (DPO: 20210114153645 register AP-HP). The study is also registered as a clinical trial.\footnote{\url{https://clinicaltrials.gov/study/NCT05089799}} Additional information can be found in~\cite{spring-d12}. In order to guarantee universal healthcare access, the Ethical Approval restricted the experiments in the hospital to an auxiliary room, instead of the main waiting room, in a way that patients unwilling to interact with the robot would not feel aside or unwelcome. \addnote[consent]{1}{This implies that only the participants who signed the informed consent were seen and heard by the robot, and data from no other person is included in the recordings or experiments}. The auxiliary room is small ($< 20$ m$^2$), \addnote[meaningless]{1}{making the robot self-localisation and navigation tasks trivial}. Hence, both skills have been validated in our respective laboratories and when possible in the Broca hospital without patients.

The Ethical Approval also restricted sharing the data collected in the hospital strictly to the partners of the SPRING project, and therefore this data can be neither shared publicly nor shared individually with anyone outside the project. The pipeline used to transfer data within the partners of the project was validated by Inria's DPO and Chief Security Officer. However, all the reports of the project describing the main findings are publicly available,\footnote{\url{https://spring-h2020.eu/results/}} as well as the code of the software modules.\footnote{\url{https://gitlab.inria.fr/spring}}

\addnote[ethics]{1}{Summarising, for this study involving an experimental product, we took care of respecting the rights and dignity of users through two main measures:
\begin{itemize}
\item\textbf{Informed consent}: Before taking part in the experiments, all users were asked to sign consent forms and image release forms.
\item\textbf{Anonymity and data protection}: The experiments took place in a closed room, which ensured that no participant who had not signed the consent form was filmed. All the videos recorded by the robot and any non-anonymous data were carefully processed and deleted, thus protecting users' privacy. This is a manual process done by the experimenter, previous to sharing the data with any of the project's partners.
\end{itemize}}

\paragraph{Computational and Storage Resources}

The experiments were conducted using the ARI robotic platform and the software architecture described in Section~\ref{sec:architecture}, as well as a dedicated external computing server.
The external computing server has one NVIDIA RTX A6000 GPU with 48 GB of VRAM. 
Additionally, a dedicated secured 4G mobile connection is used for an internet connection with a remote server running the LLM of the conversational system (Section~\ref{sub:nlu}). 
This remote server has four NVIDIA GeForce RTX 2080 Ti GPU. 
\addnote[computing]{1}{Summarising, the computations involving heavy data flows (images, audio) are run on an external computer connected to ARI via Wi-Fi, while the dialogue, requiring only textual information, is run on the server of one of the project's partners. The overall computing pipeline is privacy-preserving while ensuring enough reactivity on ARI's side.}
The collected experimental data is securely stored on a NAS server from Synology, after pseudonymisation. 
\addnote[server]{1}{The server was set up following the instructions of Inria's Chief Security Officer (in a locked closet, encrypted, and password protected).}
An experimenter tablet was available to monitor the status of the robot, using the described experimental interface (Section~\ref{sub:interface}), as well as for the experimenter to stop the robot and interaction if needed. 
An external camera was used to record the overall scene for posterior annotation and understanding.

\subsection{Performance Measures}\label{sub:expMetrics}

In this study, the user experience was assessed by two standardised scales. 
The first quantitative measure is the acceptability of the robot assessed with the Acceptability E-scale (AES)~\cite{micoulaud2016validation}.
It is designed to measure the subjective acceptability of a system using 6 items, resulting in a global score that ranges from 6 to 30. 
For consumer ready-to-use products, the acceptability cut-off score is 25/30. 
The second quantitative measure is the usability of the robot assessed with the System Usability Scale (SUS)~\cite{brooke1996system}.
It is a 10-items scale designed to assess the overall user-friendliness of a system, and generate an overall score out of 100, where a higher score indicates better user-friendliness. 
For consumer ready-to-use products, the usability cut-off score is 72/100.  
Statistical differences were assessed using independent Student t-tests.
The details of the AES and SUS scales can be found in Appendix~\ref{sec:aes-sus}.
\addnote[interviews]{1}{In addition to these two quantitative measures, we have also conducted a semi-structured interview on ethical issues, containing six open questions, ranging from the robot design to the potential emotional attachment. Other strategies could have consisted in using the story dialogue method~\cite{saplacan2023health}.
The questions and sample answers will be presented in the results section.}

 \subsection{Procedure}\label{sub:procedure}

 The participant recruitment process began two weeks before the day of the experiment. 
 A researcher used the management software of the day-care hospital~\cite{orinel2021dossier} to identify patients who met the eligibility criteria. 
 Once this information was validated, the researcher contacted the participant by telephone to present the objectives of the study and to arrange a time slot on the day of their medical visit to the hospital if the participant was willing to accept to take part in the tests. 
 The researcher sent afterwards the information letter detailing the study by postal mail, together with a reminder letter of the appointment.

On the day of the test, the researcher checked that the tools (robot, tablet, camera) %, 
were working properly. 
Once the participants arrived, the researcher reminded them of the objectives of the study and the evaluation procedures. 
Participants could be escorted by one accompanying person \addnote[together]{1}{that could participate in the experiment together with the patient. In some cases, accompanying adults were also invited to take part in the experiment on their own during their companions' appointments}. 
In case they confirmed their agreement to take part in the experiment, every participant was given a consent form to sign.
The participant(s) were then invited to stand in front of the robot to begin the interaction with it \addnote[asr-french2]{1}{in French}. 
Two researchers led the activity, a facilitator, and a person who controlled the tablet and completed the observation grid. 
The facilitator's role was to direct the test session and provide assistance or additional information to participants when needed. 
\addnote[duration]{1}{Overall, the entire interaction with one participant lasted between five and ten minutes, depending on the willingness of the participant to pursue discussing with ARI.}
%As part of the experiments at the day hospital, five scenarios were designed to improve the patient experience: a warm welcome, social interaction without health risks, help in preparing for consultations, orientation and guidance, and therapeutic entertainment. % \cite{addlesee2023data} <- paper with pictograms

\begin{figure}
    \centering
    \includegraphics[width=\columnwidth]{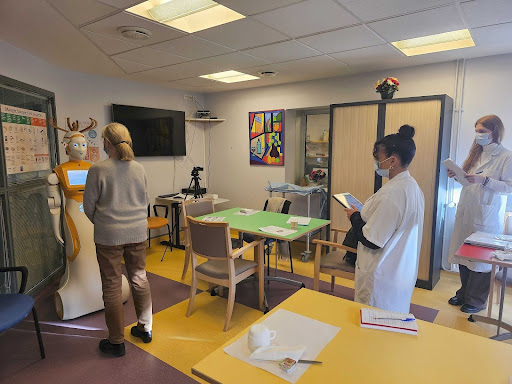}
    \caption{Experimental set up at the Broca day-care gerontology hospital. Two experimenters monitor the interaction of a patient with the ARI robot. Everything is recorded both by ARI's sensors and by the external camera.}\label{fig:setup}
\end{figure}

\addnote[documented]{1}{The experiments were documented using several devices. The robot's onboard cameras and microphones captured the interactions from an ego-centric point of view, while an external camera (1 meter away from the robot) filmed the entire scene from an external perspective for further analysis. During the interaction with the robot, the experimenter took photos of the exchange via a laboratory camera. The interviews were recorded with a dictaphone by the experimenter, who then manually transcribed the conversations, ensuring a faithful and accurate transcription for qualitative data analysis. For all the experiments, we obtained $1, 142$ minutes of interviews excluding interaction with the robot, representing $418$ pages of transcriptions.
The data was then analyzed by three laboratory researchers, who dematerialized the interviews using the Dedoose\footnote{\url{https://www.dedoose.com/}} software. Figure~\ref{fig:setup} shows the experimental setup in the day-care hospital.}

\subsection{Experimental Waves}\label{sub:waves}
\newcommand{\first}{$1^{\text{st}}$}
\newcommand{\second}{$2^{\text{nd}}$}
\newcommand{\third}{$3^{\text{rd}}$}
Two separate waves of experimentation were conducted during this study, depending on the status of the architecture.
\addnote[waves]{1}{The experiments within the same wave used the exact same software architecture, while experiments in different waves correspond to changes in some of the modules. 
The architecture hence evolved following the feedback the users provided, which was exploited to improve certain software modules as detailed below.}
%The \textit{Pre-development} wave (October 2022 -- April 2023) was used to collect data and to conduct first tests to see how patients and personnel reacted to the robot. 
The first wave (\first: May 2023 -- July 2023) used the initial version of the robot dialogue module, with an off-the-shelf automatic speech recognition model, and with almost no multi-party capabilities. 
In the second wave (\second: Sept. 2023 -- Jan. 2024), some improvements to the system were made: 
The dialogue management included large-language models, the ASR was fine-tuned, the experimenter interface was updated, all-in-all to improve the user experience. 

\subsection{Results}\label{sub:results}

% \begin{table}[t]
% \resizebox{\columnwidth}{!}{\begin{tabular}{ccccc}
% \toprule
% Wave & \multicolumn{2}{c}{\textit{First}} & \multicolumn{2}{c}{\textit{Second}} \\ 
% Variable & Count & Age & Count & Age \\\midrule
% Patients (F/M) & 15 (10/5) & 79.2(±6.62) & 33 (14/9) & 78.6 (±8.08) \\
% Companions (F/M) & 5 (4/1) & 69.2 (±15.82) & 10 (6/4) & 56.7 (±19.43) \\
% Overall (F/M) & 20 (14/6) & 76.7 (±10.23) & 43 (20/13) & 73.5 (±14.39) \\
%  \bottomrule
% %  & Patients    & Companions   & Patients    & Companions  \\ \hline
% % Count  & - & 15 & 5  &  33  & 1  \\
% % \multirow{2}{*}{Gender} & Male  & 5    & 1    &  9  & 4 \\ & Female   & 10  & 4 &  14 & 6  \\
% % Age   & \begin{tabular}[c]{@{}c@{}}Mean  Min Max\end{tabular}& 79.2 & 69.2 &  78,6  & 56,7       \\ \hline
% \end{tabular}}\vspace{2mm}
% \caption{Demographic data of participants.}
% \label{tab:demographics}
% \end{table}

\begin{table}[t]
\resizebox{\columnwidth}{!}{\begin{tabular}{clccc}
\toprule
% Wave & \multicolumn{2}{c}{\textit{First}} & \multicolumn{2}{c}{\textit{Second}} \\ 
Wave & User group & Count (F/M) & Age \\\midrule
\multirow{3}{*}{\first} & Patients & 15 (10/\phantom{0}5) & 79.2 (±\phantom{0}6.62) \\
 & Companions & \phantom{0}5 (\phantom{0}4/\phantom{0}1) & 69.2 (±15.82) \\
 & Overall & 20 (14/\phantom{0}6) & 76.7 (±10.23) \\\midrule
\multirow{3}{*}{\second} & Patients & 33 (14/\phantom{0}9) & 78.6 (±\phantom{0}8.08) \\
 & Companions & 10 (\phantom{0}6/\phantom{0}4) & 56.7 (±19.43) \\
 & Overall & 43 (20/13) & 73.5 (±14.39) \\%\midrule 
 % \multirow{3}{*}{\third} & Patients &  &  \\
 % & Companions &  &  \\
 % & Overall &  &  \\
 \bottomrule
%  & Patients    & Companions   & Patients    & Companions  \\ \hline
% Count  & - & 15 & 5  &  33  & 1  \\
% \multirow{2}{*}{Gender} & Male  & 5    & 1    &  9  & 4 \\ & Female   & 10  & 4 &  14 & 6  \\
% Age   & \begin{tabular}[c]{@{}c@{}}Mean  Min Max\end{tabular}& 79.2 & 69.2 &  78,6  & 56,7       \\ \hline
\end{tabular}}\vspace{2mm}
\caption{Demographic data of participants.}
\label{tab:demographics}
\end{table}

% \begin{table}[t]
% \resizebox{\columnwidth}{!}{\begin{tabular}{ccccc}
% \toprule
% Wave & \multicolumn{2}{c}{\textit{Initial}} & \multicolumn{2}{c}{\textit{Enhanced}} \\ 
% Variable & Count & Age & Count & Age \\\midrule
% Patients (F/M) & 15 (10/5) & 79.2 & 33 (14/9) & 78.6 \\
% Companions (F/M) & 5 (4/1) & 69.2 & 10 (6/4) & 56.7 \\
% Overall (F/M) & 20 (14/6) & 76.7 & 43 (20/13) & 73.5 \\
%  \bottomrule
% %  & Patients    & Companions   & Patients    & Companions  \\ \hline
% % Count  & - & 15 & 5  &  33  & 1  \\
% % \multirow{2}{*}{Gender} & Male  & 5    & 1    &  9  & 4 \\ & Female   & 10  & 4 &  14 & 6  \\
% % Age   & \begin{tabular}[c]{@{}c@{}}Mean  Min Max\end{tabular}& 79.2 & 69.2 &  78,6  & 56,7       \\ \hline
% \end{tabular}}\vspace{2mm}
% \caption{Socio-demographic data of the participants.}
% \label{tab:demographics}
% \end{table}

The recruitment efforts previous to each experimental wave lead to three sets of participants with slightly different age profiles for the patients, and wider age differences for the companions (see Table~\ref{tab:demographics}). The study counts a total of 63 participants with an overall average age of 74.5 (±12.31) years. 
% 
% 
% \begin{table*}
% \begin{tabular}{|cc|cc|cc|}
% \hline
% \multicolumn{2}{|c|}{Variables}                                                           & \multicolumn{2}{c}{Wave 1} & \multicolumn{2}{|c|}{Wave 2} \\ 
%                         &  & Patients    & Companions   & Patients    & Companions  \\ \hline
% Count  & - & 15 & 5  &  33  & 1  \\
% \multirow{2}{*}{Gender} & Male  & 5    & 1    &  9  & 4 \\ & Female   & 10  & 4 &  14 & 6  \\
% Age   & \begin{tabular}[c]{@{}c@{}}Mean  Min Max\end{tabular}& 79.2 & 69.2 &  78,6  & 56,7       \\ \hline
% \end{tabular}
% \caption{Socio-demographic data of participants.}
% \label{tab:demographics}
% \end{table*}
% 
We also report the average and standard deviation of the AES and SUS scores (Table~\ref{tab:results-aes-sus}), split by wave and user group, as well as the overall score per wave. \addnote[cronbach]{1}{In addition, we have computed the Cronbach's alpha coefficient for the two scales, resulting in $\alpha=0.87$ and $\alpha=0.86$ for the AES and SUS scales respectively, thus indicating internal consistency of the two measurement scales.}
% The quantitative results -- meaning the average AES and SUS scores -- obtained after the patients and companions of the two waves interacted with ARI, are shown in Table~\ref{tab:results-aes-sus}. 
% We have also included the overall scores including patients and companions.

% \begin{table}[t]
% \resizebox{\columnwidth}{!}{\begin{tabular}{ccccc}
% \toprule
% Metric & \multicolumn{2}{c}{AES} & \multicolumn{2}{c}{SUS} \\ 
% Wave & \textit{First}  & \textit{Second}   & \textit{First}    & \textit{Second}        \\ 
% \midrule
% Patients & 14.73(±5.73) & 20.65 (±6.25) & 45.5 (±20.21) & 56.83 (±12.63) \\
% Companions & 18 (±4.64) & 21 (±3.30) & 55 (±28.28) &  57.5 (±10.40)\\
% Overall & 15.46 (±5.88) & 20.75 (±5.20) & 47.86 (±24.18) & 57 (±22.88) \\
% \bottomrule
% % AES (out of 30)  & Cut Off (25) & 14,7 & 18  &  20,6  & 21
% %   \\
% % SUS (out of 100) & Cut Off (72)  & 45,4  & 55  &  56,8 & 57,5  \\ \hline
% \end{tabular}}\vspace{2mm}
% \caption{Average AES and SUS scores for the two experimental waves. The acceptability and usability scores improved from the first to the second waves, especially for patients.}
% \label{tab:results-aes-sus}
% \end{table}

\begin{table}[t]
\resizebox{\columnwidth}{!}{
\begin{tabular}{clccc}
\toprule
 % & \multicolumn{2}{c}{AES} & \multicolumn{2}{c}{SUS} \\ 
Wave & User Group & AES  &  SUS\\ 
\midrule
\multirow{3}{*}{\first} & Patients & 14.7 (±\phantom{0}5.73) & 45.5 (±20.21)  \\
& Companions & 18.0 (±\phantom{0}4.64) & 55.0 (±28.28) \\
& Overall & 15.5 (±\phantom{0}5.88) & 47.9 (±24.18) \\
\midrule
\multirow{3}{*}{\second} & Patients & 20.7 (±\phantom{0}6.25) & 56.8 (±12.63) \\
& Companions & 21.0 (±\phantom{0}3.30) &  57.5 (±10.40)\\
& Overall & 20.8 (±\phantom{0}5.20) & 57.0 (±22.88) \\
% \midrule
% \multirow{3}{*}{\third} & Patients \\
% & Companions \\
% & Overall & \\
\bottomrule
% AES (out of 30)  & Cut Off (25) & 14,7 & 18  &  20,6  & 21
%   \\
% SUS (out of 100) & Cut Off (72)  & 45,4  & 55  &  56,8 & 57,5  \\ \hline
\end{tabular}
}\vspace{2mm}
\caption{Average AES and SUS scores and their standard deviation for the two experimental waves. The acceptability and usability scores improved from the first to the second waves, especially for patients.}
\label{tab:results-aes-sus}
\end{table}

\begin{figure}[t]
\centering
\includegraphics[width=.8\columnwidth]{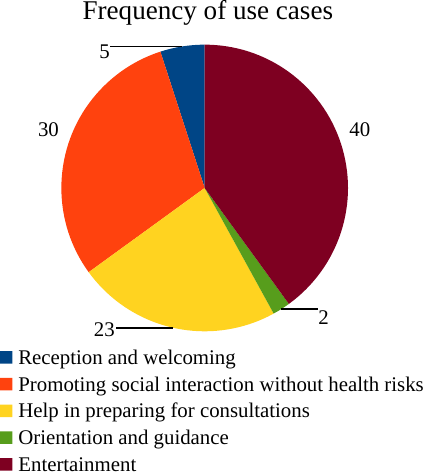}\vspace{2mm}
\caption{Proportions of use-case preferred choice.}\label{fig:use-cases}
\end{figure}

We also asked them what ``use-case'' they think would be more useful (Figure~\ref{fig:use-cases}), among the following choices: (i) reception and welcoming, (ii) promoting social interaction without health risks, (iii) help in preparing for consultations, (iv) orientation and guidance, and (v) entertainment. 
% The set of choices is a combination of the ethnographic study conducted at the beginning of the SPRING project and the initial plans submitted to the European Commission. %The choice of use-case was not conditioning anything in the system, beyond what the users themselves would say to the robot.

\begin{table*}
% \color{darkgreen}
\caption{Verbatim answers to the questions of the semi-structured interview on ethical concerns, structured by the experimental waves.}\label{tab:ethics-verbatim}
% \small
    \begin{tabular}{m{0.97\textwidth}}
        \toprule
        \textbf{Q1. Do you approve of robots that look like humans?} \vspace{1mm}\\
        ``The design is a bit too mechanical, it lacks human warmth.''\\
        % The robot isn't very visually engaging. \\
        ``I like the futuristic style, it fits with my idea of robots.''\\
        ``The design is clean, but it could be more user-friendly.''\\
        ``It's more fun and more attractive that the robot has a humanoid appearance.''\\
        ``The robot shouldn't be too realistic. It should not be an identical replica of the human, especially if it is to be used in public places.''\\
        \midrule
        \textbf{Q2. What do you think of the use of social robots in hospitals?}\vspace{1mm}\\
        ``I don't think it's really suitable for hospitals.'' \\
        ``It could be useful, but it needs to be better integrated into workflows.'' \\
        ``I think this robot could really lighten some tasks.'' \\ 
        % The system really helps save time on small tasks.\\
        ``It must not be to replace the human being. A robot that comes to help a caregiver lift a person who has fallen, I agree with that, but a robot that comes to do the consultation... no way.''\\
        ``The robot can be a precious aid, of help and assistance, but it should be used in moderation.''\\
        \midrule
        \textbf{Q3. Do you think it's right/just for robots to record and store information?} \vspace{1mm}\\
        ``I'm not comfortable with the idea of my data being stored.''\\ 
        % I think there needs to be more clarity about how data is used. \\
        ``I'm not sure what the robot is doing with my information.''\\ 
        ``I'd like to know exactly who has access to my data.''\\
        ``If it's collected online, no. If it's protected, like with computers, it's fine.''\\
        ``Data is anyway collected whether it is recorded by a robot or by a computer in the hospital.''\\
        \midrule
        \textbf{Q4. Are you concerned about the mistakes social robots could make?} \vspace{1mm}\\
        ``If the robot makes a mistake, it can have serious consequences, especially in a hospital environment.'' \\
        ``It's a bit worrying because a simple mistake could disorient someone.'' \\
        ``The robot made a mistake, but it made up for it quite quickly. That's reassuring.'' \\
        % I'm not comfortable with the idea of depending on a robot that could make mistakes.\\
        ``It can happen indeed [that the robot makes mistakes], but like any software, or any computer system, we must not worry about everything, we must progress.'' \\
        ``If the robot doesn't provide the right direction, it causes delay and irritation. But it's not like a medical intervention where it would be really dramatic.'' \\
        \midrule
        \textbf{Q5. Do you think it's possible to have an emotional bond with a robot?} \vspace{1mm}\\
        % It's a bit disturbing, you could get attached to this machine. \\
        ``I think this robot could create a kind of emotional dependency.'' \\
        ``I wonder if it's healthy to get attached to a robot.'' \\ 
        ``We could easily feel too close to the robot, that's a worry.'' \\
        ``The emotional bond can only develop from one living person to another living person. For there to be affection, there has to be a relationship, you have to be able to touch, you have to be able to feel.'' \\
        ``I don't think so, because for me it [the robot] is a machine.'' \\
        \midrule
        \textbf{Q6. Are you concerned about social robots replacing human beings?} \vspace{1mm}\\
        ``It could replace humans in certain tasks.'' \\
        % I'm worried about robots taking people's jobs. \\
        ``It's scary to think that this could replace humans.'' \\
        ``I think some jobs will be threatened by this kind of technology.''\\
        ``From a social point of view, yes, the risk of replacement is a bit worrying. Robots are useful in industry, to accomplish repetitive tasks and why not in a hospital. Then the staff at the reception desk and other staff will be unemployed if they are replaced by robots.''\\
        ``Robots cannot replace people, they can only help. It's not the same thing. A machine will never replace a human, it's impossible.''\\
        \bottomrule
    \end{tabular}
\end{table*}

Finally, Table~\ref{tab:ethics-verbatim} reports que questions of the semi-structured ethical interview, and two verbatim answers for each question, at each experimental wave. These answers were provided by patients after having completed the interaction experiment with ARI. \addnote[verbatim]{2}{The verbatim answers were analyzed following an inductive thematic analysis~\cite{beaugrand1988demarche}. Through careful reading, it consists of identifying, analysing, and reporting themes within the data, with the aim of capturing useful insights about the data in relation to our research objectives. We chose to use this method because it offers several advantages such as being very flexible (not bound to a specific theoretical framework compared to the deductive approach for instance) and being more data-oriented as the themes emerge directly from the data. Thus, we can avoid any possible bias such as the researcher's possible preconceptions. Furthermore, as our experiments took place in real-life settings with a maturing system (a social robot within a hospital), we chose an open-minded approach in order not to miss anything important.}

% \begin{table*}
% \begin{tabular}{|cc|cc|cc|}
% \hline
% \multicolumn{2}{|c|}{Variables} 
% & \multicolumn{2}{c}{Wave 1} 
% & \multicolumn{2}{|c|}{Wave 2} 
%  \\ 
%                         &                                   & Patients  & Companions   & Patients    & Companions        \\ 
% \hline
% AES (out of 30)  & Cut Off (25) & 14,7 & 18  &  20,6  & 21
%   \\
% SUS (out of 100) & Cut Off (72)  & 45,4  & 55  &  56,8 & 57,5  \\ \hline
% \end{tabular}
% \caption{Titre.}
% \label{tab:demographics2}
% \end{table*}

\subsection{Discussion}\label{sub:discussion}
% By looking at the main results, when we report the average AES and SUS results (Table~\ref{tab:results-aes-sus}), we observe a clear improvement between the First and Second wave, independently of the metric and for both the patient and companion group. 
% A significant difference was found for the AES 
% ($t(20)$\hspace{0.5mm}$=$\hspace{0.5mm}$-2.991$, $p$\hspace{0.5mm}$<$\hspace{0.5mm}$.05$) 
% as the scores increased between the first ($M$\hspace{0.5mm}$=$\hspace{0.5mm}$15.5$, \textit{SD}\hspace{0.5mm}$=$\hspace{0.5mm}$5.88$) and the second wave 
% ($M$\hspace{0.5mm}$=$\hspace{0.5mm}$20.8$, \textit{SD}\hspace{0.5mm}$=$\hspace{0.5mm}$5.20$). 
% This significant difference was only found in patients ($t(18)$\hspace{0.5mm}$=$\hspace{0.5mm}$-2.906$, $p$\hspace{0.5mm}$<$\hspace{0.5mm}$.05$) and not in companions. 
% However, no significant differences were observed for the SUS between the first and the second wave of experiments, both in patients and in companions.
% While there is an overall clear trend of improvement, the trend for the companion group is less pronounced than for the patient group.

\addnote[anova]{1}{The main results, as summarized by the average AES and SUS scores (Table~\ref{tab:results-aes-sus}), reveal a clear improvement from the first to the second wave, regardless of the metric used or the participant group (patient or companion). A significant increase in AES scores was observed, based on a two-way ANOVA analysis with experimental waves and participant profiles as variables. 
Specifically, AES scores rose significantly from the first wave ($M = 15.5$, $SD = 5.88$) to the second wave ($M = 20.8$, $SD = 5.20$), $F(1,52) = 10.42$, $p < .001$, $\eta^2 = .17$. %, as revealed by Tukey's post hoc test. \todo{waiting for Sebastien}
However, this significant difference was only present among patients ($t(18)=-2.906$, $p<.05$) and not among companions.
No significant differences were found in SUS scores between the first and second waves for either group. Overall, there is a noticeable trend of improvement, although it is more pronounced among patients than among companions.}

We explain this improvement in acceptability and usability by the technical improvements we made to the software architecture (since the robot's appearance did not change between waves): namely, the fine-tuning of the ASR module and the inclusion of LLMs in the conversation manager. These two major modifications, together with bug fixes, allowed a more natural interaction with the users since (i) ARI understood better what the participants were saying and (ii) ARI was able to answer to a wider range of questions with reasonable (although now always exact) answers. In particular, the participant's feedback reported: enjoyment in using the robot, usefulness, acceptability of the robot's reaction time (talking), and overall satisfaction with the robot.

It is also interesting to observe, when considering the two groups separately, that the companions tend to provide more positive AES and SUS feedback than the patients. For the time being, we have not identified a reason that justifies this difference.

\addnote[discussion-ethics]{1}{Regarding the semi-structured interview dealing with ethical issues (see verbatim excerpts in Table~\ref{tab:ethics-verbatim}), we extract the following main messages summarising our experience. Generally speaking, participants found that ARI's design was acceptable but could be improved, in the sense that it looked a bit too mechanical and would benefit from a warmer look. The robot's usefulness was globally acknowledged, but it was also pointed out that there was room for improvement in terms of integration in the workflow (we did not target such integration in our experiments). Data use and privacy are major concerns, and participants expressed their need for transparency in terms of data use, sharing, storage, etc. We believe this awareness demonstrates that older adults are well informed about the dangers of misuse of personal data, and we interpret this request for transparency as a very good societal sign. We also reiterate that for our experiments the data was not shared beyond the SPRING consortium, in agreement with the Ethical Approvals and the protocol validated by Inria's Chief Security Officer. The intertwining between the first and third questions (appearance and privacy) appears to be a complex matter~\cite{lindblom2024qualitative}, since users link certain appearance features to their privacy being or not preserved by the robot. 
Regarding the potential mistakes made by the robot, the participants were moderately concerned but felt reassured by having human supervision. Participants were also aware of the risk of attachment to the robot, and raised the concern of preventing this from happening, especially with vulnerable persons. Finally, the participants were significantly concerned about the potential replacement of human jobs by robots, pointing out that they would be useful for specific small tasks. Overall, these verbatim answers provided an interesting view of how older adults perceive the use of socially assistive robots in healthcare facilities, and they are very complementary to the quantitative acceptability and usability scales results discussed above.
In addition, some of the ethical concerns expressed by the participants (loss of privacy, decreased human contact/human replacement) are aligned with existing ethical discussions in the literature~\cite{saplacan2021ethical}, while others (e.g.\ developing an emotional atttachment) are complementary.}
% The discussions surrounding the experiments.

\addnote[novelty-discussion]{1}{All these experiments were possible thanks to the software architecture described in Section~\ref{sec:architecture} (as well as to the designed experimental protocol). To our knowledge, this is the first robotic software architecture enabling audio-visual interaction of a robotic platform with multiple persons. It is also one of the first architectures exploiting the ROS4HRI standard for human-robot interaction. Very importantly, this is the first time such an advanced interaction architecture has been validated with more than 60 experiments with real patients and companions in a day-care gerontological hospital.}

\subsection{Challenges and Failure Cases}\label{sub:failures}

One of the major challenges we encountered during our project concerned the recruitment of patients for the tests. These challenges can be attributed to a number of factors, including organisational aspects, last-minute cancellations and refusal to interact with the robot.

Some factors had a significant impact on the organisation of the experiments, for example, patients that refused to take part in the experiments, and coordination difficulties among the various hospital stakeholders (patients arriving late, consultations' delay, or some experiments lasting longer than planned resulting in a delay in the following appointments). Finally, some participants refused to interact with the ARI robot during the first physical encounter, which directly compromised the interaction and made the experiment impossible. The reasons for this refusal were diverse, ranging from technological apprehension linked to the size of the robot, and ethical concerns.

During the First wave, some participants felt frustration specifically due to the limitations in terms of conversational capabilities, whether these were related to the performance of the ASR or the topic restrictions of the dialogue manager. Since these were identified as the two main blocking points, we paid attention to enhancing these specific skills, leading to a clear improvement in ARI's acceptability and usability in the Second wave.

% \note{What went wrong, what did the patients, companions, and experimentors find frustrating. What was changed to address these failures in current system.}

\section{Conclusion and Open Questions}\label{sec:conc}
% \note{Conclusion including open topics.}

In this paper, we have investigated the acceptability and usability of a social robot in gerontology healthcare. Compared to previous research~\cite{pedersen2018developing,gongora2019social}, our study is a step forward mainly because of three reasons. First, the acceptability and usability is evaluated by patients and companions within their regular visits to a day-care hospital, which is different to the more common scenarios of nursing facilities and private homes. Second, the platform used is a full-sized humanoid robot, very different in size and appearance from pet-like and small-sized humanoid robots. Finally, the platform enabled multi-modal conversational interaction, which is again uncommon in many previous studies. The combination of these three different characteristics makes this study the first of its kind, and we hope it opens the door to multiple follow-ups and a wider evaluation.

The paper describes the overall robotic and software architecture and provides details of the various modules and methods used for the experiments. We also discussed the materials, methods, and recruitment process, and provided technical and experimental details. After two experimental waves, we can provide an assessment in terms of the acceptability and usability of the developed technology. The most important result, of such human-robot interaction experiments, is that the improvements (ASR robustness, dialogue flexibility) had a positive effect on how the system is perceived by the patients and companions in the Broca gerontology day-care  hospital.

The study and associated technology present several limitations. First, given that all participants were requested to sign a consent form, the robot was never facing people unwilling to interact, and therefore we were not able to test its ability to properly understand the lack of interaction interest and execute consequent actions (e.g.\ leaving the person alone). Second, from a technical perspective, the experiments require dedicated computational power, which might limit the deployment of such technology. The question of how to provide state-of-the-art perception and action skills for a social robot with limited on-board computational resources is widely open, and not easy to address. Third, other social skills such as the ability to hold conversations within groups (multi-party dialogue) or navigating while accounting for the presence of humans (social navigation) were not evaluated in this study. Fourth, it would be interesting to run the same evaluation with medical personnel and understand if there are important perception differences in terms of acceptability and usability with respect to patients and companions.
Finally, beyond social skills, it would be interesting to evaluate the capacity of the robot to be connected to the information system of the hospital for logistic purposes (e.g.\ reminding appointments, rescheduling them, providing information about the doctor's office or name), but this poses important ethical and security issues that have to be very carefully addressed. The proper evaluation of how these capabilities are seen and welcomed by the patients, companions and medical personnel is crucial to understand its impact on the everyday life of the hospital.

\backmatter

% \bmhead{Supplementary information}

% If your article has accompanying supplementary file/s please state so here.

% Authors reporting data from electrophoretic gels and blots should supply the full unprocessed scans for key as part of their Supplementary information. This may be requested by the editorial team/s if it is missing.

% Please refer to Journal-level guidance for any specific requirements.

\section*{Declarations}
\subsection*{Ethics Statement}
As detailed in Section~\ref{sub:protocol}, the study was approved by the French
National Ethical Review Board (\textit{Comité National de Protection des Personnes, CPP Ouest II,
Maison de la Recherche Clinique-CHU Angers}, approval number: 2021/20), the local ethics
committee of the University of Paris (\textit{Comité d'Ethique de la Recherche} CER-N IRB: 00012020-108), and was compliant with the General Data Protection Regulation (Data Protection Officer register 20210114153645 at AP-HP). The study is also registered as
a clinical trial, see \url{https://clinicaltrials.gov/study/NCT05089799}.

\subsection*{Data Availability Statement}
The Ethical Approval restricted sharing the data collected in the hospital strictly to the
partners of the SPRING project, and therefore this data can be neither shared publicly nor shared
individually with anyone outside the project, see Section~\ref{sub:protocol}.

\section*{Acknowledgments}

This research was funded by the EU H2020 program under grant agreement no. 871245 (\url{https://spring-h2020.eu/}). We would also like to thank our anonymous reviewers for their time and valuable feedback.

\section*{Author contributions}
The complete list of author contributions is detailed in Table~\ref{tab:authors} at the end of the document, in which we report the participation of each author according to the items in the CRediT taxonomy.\footnote{\url{https://credit.niso.org/}} The first four authors contributed equally to the paper, being the first one the Coordinator of the SPRING project and corresponding author, the other three are ordered alphabetically. The rest of the authors are also ordered alphabetically. 

% \note{Author contributions are described following the CRediT taxonomy (\url{https://credit.niso.org/}).}
% The following roles exist: 
% \begin{itemize}
%     \item Conceptualization
%     \item Data curation
%     \item Formal Analysis
%     \item Funding acquisition
%     \item Investigation
%     \item Methodology
%     \item Project administration
%     \item Resources
%     \item Software
%     \item Supervision
%     \item Validation
%     \item Visualization
%     \item Writing – original draft
%     \item Writing – review \& editing
% \end{itemize}
% \note{Maybe add a table where writing and software is attributed by section.}

\bibliography{sn-bibliography}% common bib file


\begin{thebibliography}{97}
\ifx \bisbn   \undefined \def \bisbn  #1{ISBN #1}\fi
\ifx \binits  \undefined \def \binits#1{#1}\fi
\ifx \bauthor  \undefined \def \bauthor#1{#1}\fi
\ifx \batitle  \undefined \def \batitle#1{#1}\fi
\ifx \bjtitle  \undefined \def \bjtitle#1{#1}\fi
\ifx \bvolume  \undefined \def \bvolume#1{\textbf{#1}}\fi
\ifx \byear  \undefined \def \byear#1{#1}\fi
\ifx \bissue  \undefined \def \bissue#1{#1}\fi
\ifx \bfpage  \undefined \def \bfpage#1{#1}\fi
\ifx \blpage  \undefined \def \blpage #1{#1}\fi
\ifx \burl  \undefined \def \burl#1{\textsf{#1}}\fi
\ifx \doiurl  \undefined \def \doiurl#1{\url{https://doi.org/#1}}\fi
\ifx \betal  \undefined \def \betal{\textit{et al.}}\fi
\ifx \binstitute  \undefined \def \binstitute#1{#1}\fi
\ifx \binstitutionaled  \undefined \def \binstitutionaled#1{#1}\fi
\ifx \bctitle  \undefined \def \bctitle#1{#1}\fi
\ifx \beditor  \undefined \def \beditor#1{#1}\fi
\ifx \bpublisher  \undefined \def \bpublisher#1{#1}\fi
\ifx \bbtitle  \undefined \def \bbtitle#1{#1}\fi
\ifx \bedition  \undefined \def \bedition#1{#1}\fi
\ifx \bseriesno  \undefined \def \bseriesno#1{#1}\fi
\ifx \blocation  \undefined \def \blocation#1{#1}\fi
\ifx \bsertitle  \undefined \def \bsertitle#1{#1}\fi
\ifx \bsnm \undefined \def \bsnm#1{#1}\fi
\ifx \bsuffix \undefined \def \bsuffix#1{#1}\fi
\ifx \bparticle \undefined \def \bparticle#1{#1}\fi
\ifx \barticle \undefined \def \barticle#1{#1}\fi
\bibcommenthead
\ifx \bconfdate \undefined \def \bconfdate #1{#1}\fi
\ifx \botherref \undefined \def \botherref #1{#1}\fi
\ifx \url \undefined \def \url#1{\textsf{#1}}\fi
\ifx \bchapter \undefined \def \bchapter#1{#1}\fi
\ifx \bbook \undefined \def \bbook#1{#1}\fi
\ifx \bcomment \undefined \def \bcomment#1{#1}\fi
\ifx \oauthor \undefined \def \oauthor#1{#1}\fi
\ifx \citeauthoryear \undefined \def \citeauthoryear#1{#1}\fi
\ifx \endbibitem  \undefined \def \endbibitem {}\fi
\ifx \bconflocation  \undefined \def \bconflocation#1{#1}\fi
\ifx \arxivurl  \undefined \def \arxivurl#1{\textsf{#1}}\fi
\csname PreBibitemsHook\endcsname

%%% 1
\bibitem[\protect\citeauthoryear{Leite et~al.}{2013}]{leite2013social}
\begin{barticle}
\bauthor{\bsnm{Leite}, \binits{I.}},
\bauthor{\bsnm{Martinho}, \binits{C.}},
\bauthor{\bsnm{Paiva}, \binits{A.}}:
\batitle{Social robots for long-term interaction: a survey}.
\bjtitle{International Journal of Social Robotics}
\bvolume{5},
\bfpage{291}--\blpage{308}
(\byear{2013})
\end{barticle}
\endbibitem

%%% 2
\bibitem[\protect\citeauthoryear{Breazeal et~al.}{2016}]{breazeal2016social}
\begin{botherref}
\oauthor{\bsnm{Breazeal}, \binits{C.}},
\oauthor{\bsnm{Dautenhahn}, \binits{K.}},
\oauthor{\bsnm{Kanda}, \binits{T.}}:
Social robotics.
Springer handbook of robotics,
1935--1972
(2016)
\end{botherref}
\endbibitem

%%% 3
\bibitem[\protect\citeauthoryear{Thrun}{1998}]{thrun1998robots}
\begin{barticle}
\bauthor{\bsnm{Thrun}, \binits{S.}}:
\batitle{{When robots meet people}}.
\bjtitle{IEEE Intelligent Systems and their Applications}
\bvolume{13}(\bissue{3}),
\bfpage{27}--\blpage{29}
(\byear{1998})
\end{barticle}
\endbibitem

%%% 4
\bibitem[\protect\citeauthoryear{Al~Moubayed et~al.}{2012}]{al2012furhat}
\begin{bchapter}
\bauthor{\bsnm{Al~Moubayed}, \binits{S.}},
\bauthor{\bsnm{Beskow}, \binits{J.}},
\bauthor{\bsnm{Skantze}, \binits{G.}},
\bauthor{\bsnm{Granstr{\"o}m}, \binits{B.}}:
\bctitle{Furhat: a back-projected human-like robot head for multiparty
  human-machine interaction}.
In: \bbtitle{Cognitive Behavioural Systems: COST 2102 International Training
  School, Dresden, Germany, February 21-26, 2011, Revised Selected Papers},
pp. \bfpage{114}--\blpage{130}
(\byear{2012}).
\bcomment{Springer}
\end{bchapter}
\endbibitem

%%% 5
\bibitem[\protect\citeauthoryear{Keizer et~al.}{2014}]{keizer2014machine}
\begin{barticle}
\bauthor{\bsnm{Keizer}, \binits{S.}},
\bauthor{\bsnm{Ellen~Foster}, \binits{M.}},
\bauthor{\bsnm{Wang}, \binits{Z.}},
\bauthor{\bsnm{Lemon}, \binits{O.}}:
\batitle{Machine learning for social multiparty human--robot interaction}.
\bjtitle{ACM transactions on interactive intelligent systems (TIIS)}
\bvolume{4}(\bissue{3}),
\bfpage{1}--\blpage{32}
(\byear{2014})
\end{barticle}
\endbibitem

%%% 6
\bibitem[\protect\citeauthoryear{Furhat~Robotics}{2015}]{robotics2018franny}
\begin{botherref}
\oauthor{\bsnm{Furhat~Robotics}, \binits{P.}}:
Franny, frankfurt airport’s new multilingual robot concierge can help you in
  over 35 languages.
Furhat Robotics Press Release
(2015)
\end{botherref}
\endbibitem

%%% 7
\bibitem[\protect\citeauthoryear{Foster et~al.}{2019}]{foster2019mummer}
\begin{botherref}
\oauthor{\bsnm{Foster}, \binits{M.E.}},
\oauthor{\bsnm{Craenen}, \binits{B.}},
\oauthor{\bsnm{Deshmukh}, \binits{A.}},
\oauthor{\bsnm{Lemon}, \binits{O.}},
\oauthor{\bsnm{Bastianelli}, \binits{E.}},
\oauthor{\bsnm{Dondrup}, \binits{C.}},
\oauthor{\bsnm{Papaioannou}, \binits{I.}},
\oauthor{\bsnm{Vanzo}, \binits{A.}},
\oauthor{\bsnm{Odobez}, \binits{J.-M.}},
\oauthor{\bsnm{Can{\'e}vet}, \binits{O.}}, et al.:
{MuMMER: Socially intelligent human-robot interaction in public spaces}.
arXiv preprint arXiv:1909.06749
(2019)
\end{botherref}
\endbibitem

%%% 8
\bibitem[\protect\citeauthoryear{Vlachos et~al.}{2020}]{vlachos2020robot}
\begin{bchapter}
\bauthor{\bsnm{Vlachos}, \binits{E.}},
\bauthor{\bsnm{Hansen}, \binits{A.F.}},
\bauthor{\bsnm{Holck}, \binits{J.P.}}:
\bctitle{A robot in the library}.
In: \bbtitle{International Conference on Human-computer Interaction},
pp. \bfpage{312}--\blpage{322}
(\byear{2020}).
\bcomment{Springer}
\end{bchapter}
\endbibitem

%%% 9
\bibitem[\protect\citeauthoryear{Gunson et~al.}{2022}]{gunson2022developing}
\begin{bchapter}
\bauthor{\bsnm{Gunson}, \binits{N.}},
\bauthor{\bsnm{Garc{\'\i}a}, \binits{D.H.}},
\bauthor{\bsnm{Siei{\'n}ska}, \binits{W.}},
\bauthor{\bsnm{Dondrup}, \binits{C.}},
\bauthor{\bsnm{Lemon}, \binits{O.}}:
\bctitle{Developing a social conversational robot for the hospital waiting
  room}.
In: \bbtitle{2022 31st IEEE International Conference on Robot and Human
  Interactive Communication (RO-MAN)},
pp. \bfpage{1352}--\blpage{1357}
(\byear{2022}).
\bcomment{IEEE}
\end{bchapter}
\endbibitem

%%% 10
\bibitem[\protect\citeauthoryear{Khaksar et~al.}{2023}]{khaksar2023robotics}
\begin{botherref}
\oauthor{\bsnm{Khaksar}, \binits{W.}},
\oauthor{\bsnm{Saplacan}, \binits{D.}},
\oauthor{\bsnm{Bygrave}, \binits{L.A.}},
\oauthor{\bsnm{Torresen}, \binits{J.}}:
Robotics in elderly healthcare: a review of 20 recent research projects.
arXiv preprint arXiv:2302.04478
(2023)
\end{botherref}
\endbibitem

%%% 11
\bibitem[\protect\citeauthoryear{Sackl et~al.}{2022}]{sackl2022social}
\begin{bchapter}
\bauthor{\bsnm{Sackl}, \binits{A.}},
\bauthor{\bsnm{Pretolesi}, \binits{D.}},
\bauthor{\bsnm{Burger}, \binits{S.}},
\bauthor{\bsnm{Ganglbauer}, \binits{M.}},
\bauthor{\bsnm{Tscheligi}, \binits{M.}}:
\bctitle{{Social Robots as Coaches: How Human-Robot Interaction Positively
  Impacts Motivation in Sports Training Sessions}}.
In: \bbtitle{2022 31st IEEE International Conference on Robot and Human
  Interactive Communication (RO-MAN)},
pp. \bfpage{141}--\blpage{148}
(\byear{2022}).
\bcomment{IEEE}
\end{bchapter}
\endbibitem

%%% 12
\bibitem[\protect\citeauthoryear{Langedijk
  et~al.}{2020}]{langedijk2020studying}
\begin{bchapter}
\bauthor{\bsnm{Langedijk}, \binits{R.M.}},
\bauthor{\bsnm{Odabasi}, \binits{C.}},
\bauthor{\bsnm{Fischer}, \binits{K.}},
\bauthor{\bsnm{Graf}, \binits{B.}}:
\bctitle{{Studying drink-serving service robots in the real world}}.
In: \bbtitle{2020 29th IEEE International Conference on Robot and Human
  Interactive Communication (RO-MAN)},
pp. \bfpage{788}--\blpage{793}
(\byear{2020}).
\bcomment{IEEE}
\end{bchapter}
\endbibitem

%%% 13
\bibitem[\protect\citeauthoryear{Stegner et~al.}{2023}]{stegner2023situated}
\begin{bchapter}
\bauthor{\bsnm{Stegner}, \binits{L.}},
\bauthor{\bsnm{Senft}, \binits{E.}},
\bauthor{\bsnm{Mutlu}, \binits{B.}}:
\bctitle{{Situated participatory design: A method for in situ design of robotic
  interaction with older adults}}.
In: \bbtitle{Proceedings of the 2023 CHI Conference on Human Factors in
  Computing Systems},
pp. \bfpage{1}--\blpage{15}
(\byear{2023})
\end{bchapter}
\endbibitem

%%% 14
\bibitem[\protect\citeauthoryear{Blavette et~al.}{2022}]{blavette2022robot}
\begin{barticle}
\bauthor{\bsnm{Blavette}, \binits{L.}},
\bauthor{\bsnm{Rigaud}, \binits{A.-S.}},
\bauthor{\bsnm{Anzalone}, \binits{S.M.}},
\bauthor{\bsnm{Kergueris}, \binits{C.}},
\bauthor{\bsnm{Isabet}, \binits{B.}},
\bauthor{\bsnm{Dacunha}, \binits{S.}},
\bauthor{\bsnm{Pino}, \binits{M.}}:
\batitle{A robot-mediated activity using the nao robot to promote covid-19
  precautionary measures among older adults in geriatric facilities}.
\bjtitle{International Journal of Environmental Research and Public Health}
\bvolume{19}(\bissue{9}),
\bfpage{5222}
(\byear{2022})
\end{barticle}
\endbibitem

%%% 15
\bibitem[\protect\citeauthoryear{Gonz{\'a}lez-Gonz{\'a}lez
  et~al.}{2021}]{gonzalez2021social}
\begin{barticle}
\bauthor{\bsnm{Gonz{\'a}lez-Gonz{\'a}lez}, \binits{C.S.}},
\bauthor{\bsnm{Violant-Holz}, \binits{V.}},
\bauthor{\bsnm{Gil-Iranzo}, \binits{R.M.}}:
\batitle{Social robots in hospitals: a systematic review}.
\bjtitle{Applied Sciences}
\bvolume{11}(\bissue{13}),
\bfpage{5976}
(\byear{2021})
\end{barticle}
\endbibitem

%%% 16
\bibitem[\protect\citeauthoryear{Hahkio}{2020}]{hahkio2020service}
\begin{botherref}
\oauthor{\bsnm{Hahkio}, \binits{L.}}:
{Service robots’ feasibility in the hotel industry: A case study of Hotel
  Presidentti}.
PhD thesis,
Laurea University of Applied Sciences
(2020)
\end{botherref}
\endbibitem

%%% 17
\bibitem[\protect\citeauthoryear{Wagner et~al.}{2023}]{wagner2023comparing}
\begin{bchapter}
\bauthor{\bsnm{Wagner}, \binits{N.}},
\bauthor{\bsnm{Kraus}, \binits{M.}},
\bauthor{\bsnm{Lindemann}, \binits{N.}},
\bauthor{\bsnm{Minker}, \binits{W.}}:
\bctitle{{Comparing Multi-User Interaction Strategies in Human-Robot
  Teamwork}}.
In: \bbtitle{The International Workshop on Spoken Dialogue Systems Technology,
  IWSDS 2023},
p. \bfpage{12}
(\byear{2023})
\end{bchapter}
\endbibitem

%%% 18
\bibitem[\protect\citeauthoryear{Cooper et~al.}{2023}]{cooper2023challenges}
\begin{bchapter}
\bauthor{\bsnm{Cooper}, \binits{S.}},
\bauthor{\bsnm{Ros}, \binits{R.}},
\bauthor{\bsnm{Lemaignan}, \binits{S.}}:
\bctitle{Challenges of deploying assistive robots in real-life scenarios: an
  industrial perspective}.
In: \bbtitle{IEEE International Conference on Robot and Human Interactive
  Communication (RO-MAN)},
pp. \bfpage{435}--\blpage{442}
(\byear{2023})
\end{bchapter}
\endbibitem

%%% 19
\bibitem[\protect\citeauthoryear{Xu et~al.}{2022}]{xu2022transcenter}
\begin{barticle}
\bauthor{\bsnm{Xu}, \binits{Y.}},
\bauthor{\bsnm{Ban}, \binits{Y.}},
\bauthor{\bsnm{Delorme}, \binits{G.}},
\bauthor{\bsnm{Gan}, \binits{C.}},
\bauthor{\bsnm{Rus}, \binits{D.}},
\bauthor{\bsnm{Alameda-Pineda}, \binits{X.}}:
\batitle{Transcenter: Transformers with dense representations for
  multiple-object tracking}.
\bjtitle{IEEE Transactions on Pattern Analysis and Machine Intelligence}
\bvolume{45}(\bissue{6}),
\bfpage{7820}--\blpage{7835}
(\byear{2022})
\end{barticle}
\endbibitem

%%% 20
\bibitem[\protect\citeauthoryear{Dubenova et~al.}{2022}]{dubenova2022dinloc}
\begin{bchapter}
\bauthor{\bsnm{Dubenova}, \binits{M.}},
\bauthor{\bsnm{Zderadickova}, \binits{A.}},
\bauthor{\bsnm{Kafka}, \binits{O.}},
\bauthor{\bsnm{Pajdla}, \binits{T.}},
\bauthor{\bsnm{Polic}, \binits{M.}}:
\bctitle{D-inloc++: Indoor localization in dynamic environments}.
In: \bbtitle{Pattern Recognition},
pp. \bfpage{246}--\blpage{261}.
\bpublisher{Springer},
\blocation{Cham}
(\byear{2022})
\end{bchapter}
\endbibitem

%%% 21
\bibitem[\protect\citeauthoryear{Andriella et~al.}{2024}]{andriella2024dataset}
\begin{bchapter}
\bauthor{\bsnm{Andriella}, \binits{A.}},
\bauthor{\bsnm{Ros}, \binits{R.}},
\bauthor{\bsnm{Ellinson}, \binits{Y.}},
\bauthor{\bsnm{Gannot}, \binits{S.}},
\bauthor{\bsnm{Lemaignan}, \binits{S.}}:
\bctitle{Dataset and evaluation of automatic speech recognition for
  multi-lingual intent recognition on social robots}.
In: \bbtitle{Proceedings of the ACM International Conference on Human Robot
  Interaction}
(\byear{2024})
\end{bchapter}
\endbibitem

%%% 22
\bibitem[\protect\citeauthoryear{Tonini et~al.}{2023}]{tonini2023object}
\begin{bchapter}
\bauthor{\bsnm{Tonini}, \binits{F.}},
\bauthor{\bsnm{Dall'Asen}, \binits{N.}},
\bauthor{\bsnm{Beyan}, \binits{C.}},
\bauthor{\bsnm{Ricci}, \binits{E.}}:
\bctitle{Object-aware gaze target detection}.
In: \bbtitle{Proceedings of the IEEE/CVF International Conference on Computer
  Vision}
(\byear{2023})
\end{bchapter}
\endbibitem

%%% 23
\bibitem[\protect\citeauthoryear{Berdasco et~al.}{2019}]{berdasco2019user}
\begin{botherref}
\oauthor{\bsnm{Berdasco}, \binits{A.}},
\oauthor{\bsnm{L{\'o}pez}, \binits{G.}},
\oauthor{\bsnm{Diaz}, \binits{I.}},
\oauthor{\bsnm{Quesada}, \binits{L.}},
\oauthor{\bsnm{Guerrero}, \binits{L.A.}}:
User experience comparison of intelligent personal assistants: Alexa, google
  assistant, siri and cortana.
UCAml 2019,
51
(2019)
\end{botherref}
\endbibitem

%%% 24
\bibitem[\protect\citeauthoryear{Addlesee
  et~al.}{2024}]{addlesee2024multiparty}
\begin{bchapter}
\bauthor{\bsnm{Addlesee}, \binits{A.}},
\bauthor{\bsnm{Cherakara}, \binits{N.}},
\bauthor{\bsnm{Nelson}, \binits{N.}},
\bauthor{\bsnm{Garc\'ia}, \binits{D.H.}},
\bauthor{\bsnm{Gunson}, \binits{N.}},
\bauthor{\bsnm{Siei{\'n}ska}, \binits{W.}},
\bauthor{\bsnm{Dondrup}, \binits{C.}},
\bauthor{\bsnm{Lemon}, \binits{O.}}:
\bctitle{Multi-party multimodal conversations between patients, their
  companions, and a social robot in a hospital memory clinic}.
In: \bbtitle{Proceedings of the 18th Conference of the European Chapter of the
  Association for Computational Linguistics (EACL)}
(\byear{2024})
\end{bchapter}
\endbibitem

%%% 25
\bibitem[\protect\citeauthoryear{Pikuli et~al.}{2024}]{pikuli2024navigating}
\begin{bchapter}
\bauthor{\bsnm{Pikuli}, \binits{D.}},
\bauthor{\bsnm{Cosio}, \binits{J.}},
\bauthor{\bsnm{Alameda-Pineda}, \binits{X.}},
\bauthor{\bsnm{Wieber}, \binits{P.-B.}},
\bauthor{\bsnm{Fraichard}, \binits{T.}}:
\bctitle{Navigating the practical pitfalls of reinforcement learning for social
  robot navigation}.
In: \bbtitle{Robotics: Science and Systems (RSS) Workshop on Unsolved Problems
  in Social Robot Navigation}
(\byear{2024})
\end{bchapter}
\endbibitem

%%% 26
\bibitem[\protect\citeauthoryear{Alameda-Pineda and
  Horaud}{2015}]{alameda2015vision}
\begin{barticle}
\bauthor{\bsnm{Alameda-Pineda}, \binits{X.}},
\bauthor{\bsnm{Horaud}, \binits{R.}}:
\batitle{Vision-guided robot hearing}.
\bjtitle{The International Journal of Robotics Research}
\bvolume{34}(\bissue{4-5}),
\bfpage{437}--\blpage{456}
(\byear{2015})
\end{barticle}
\endbibitem

%%% 27
\bibitem[\protect\citeauthoryear{Mohamed and
  Lemaignan}{2021}]{mohamed2021ros4hri}
\begin{bchapter}
\bauthor{\bsnm{Mohamed}, \binits{Y.}},
\bauthor{\bsnm{Lemaignan}, \binits{S.}}:
\bctitle{Ros for human-robot interaction}.
In: \bbtitle{Proceedings of the 2021 IEEE/RSJ International Conference on
  Intelligent Robots and Systems}
(\byear{2021})
\end{bchapter}
\endbibitem

%%% 28
\bibitem[\protect\citeauthoryear{Cooper et~al.}{2020}]{Cooper2020}
\begin{bchapter}
\bauthor{\bsnm{Cooper}, \binits{S.}},
\bauthor{\bsnm{{Di Fava}}, \binits{A.}},
\bauthor{\bsnm{Vivas}, \binits{C.}},
\bauthor{\bsnm{Marchionni}, \binits{L.}},
\bauthor{\bsnm{Ferro}, \binits{F.}}:
\bctitle{{ARI: The Social Assistive Robot and Companion}}.
In: \bbtitle{29th IEEE International Conference on Robot and Human Interactive
  Communication, RO-MAN 2020},
pp. \bfpage{745}--\blpage{751}
(\byear{2020})
\end{bchapter}
\endbibitem

%%% 29
\bibitem[\protect\citeauthoryear{Labb{\'e} and Michaud}{2019}]{labbe2019rtab}
\begin{barticle}
\bauthor{\bsnm{Labb{\'e}}, \binits{M.}},
\bauthor{\bsnm{Michaud}, \binits{F.}}:
\batitle{Rtab-map as an open-source lidar and visual simultaneous localization
  and mapping library for large-scale and long-term online operation}.
\bjtitle{Journal of field robotics}
\bvolume{36}(\bissue{2}),
\bfpage{416}--\blpage{446}
(\byear{2019})
\end{barticle}
\endbibitem

%%% 30
\bibitem[\protect\citeauthoryear{Mur-Artal and Tard{\'o}s}{2017}]{mur2017orb}
\begin{barticle}
\bauthor{\bsnm{Mur-Artal}, \binits{R.}},
\bauthor{\bsnm{Tard{\'o}s}, \binits{J.D.}}:
\batitle{Orb-slam2: An open-source slam system for monocular, stereo, and rgb-d
  cameras}.
\bjtitle{IEEE transactions on robotics}
\bvolume{33}(\bissue{5}),
\bfpage{1255}--\blpage{1262}
(\byear{2017})
\end{barticle}
\endbibitem

%%% 31
\bibitem[\protect\citeauthoryear{Sarlin et~al.}{2019}]{sarlin2019coarse}
\begin{bchapter}
\bauthor{\bsnm{Sarlin}, \binits{P.-E.}},
\bauthor{\bsnm{Cadena}, \binits{C.}},
\bauthor{\bsnm{Siegwart}, \binits{R.}},
\bauthor{\bsnm{Dymczyk}, \binits{M.}}:
\bctitle{From coarse to fine: Robust hierarchical localization at large scale}.
In: \bbtitle{CVPR}
(\byear{2019})
\end{bchapter}
\endbibitem

%%% 32
\bibitem[\protect\citeauthoryear{Lee et~al.}{2023}]{lee2023revisiting}
\begin{bchapter}
\bauthor{\bsnm{Lee}, \binits{S.}},
\bauthor{\bsnm{Lee}, \binits{S.}},
\bauthor{\bsnm{Seong}, \binits{H.}},
\bauthor{\bsnm{Kim}, \binits{E.}}:
\bctitle{Revisiting self-similarity: Structural embedding for image retrieval}.
In: \bbtitle{Proceedings of the IEEE/CVF Conference on Computer Vision and
  Pattern Recognition},
pp. \bfpage{23412}--\blpage{23421}
(\byear{2023})
\end{bchapter}
\endbibitem

%%% 33
\bibitem[\protect\citeauthoryear{Kukelova et~al.}{2016}]{kukelova2016efficient}
\begin{bchapter}
\bauthor{\bsnm{Kukelova}, \binits{Z.}},
\bauthor{\bsnm{Heller}, \binits{J.}},
\bauthor{\bsnm{Fitzgibbon}, \binits{A.}}:
\bctitle{Efficient intersection of three quadrics and applications in computer
  vision}.
In: \bbtitle{Proceedings of the IEEE Conference on Computer Vision and Pattern
  Recognition},
pp. \bfpage{1799}--\blpage{1808}
(\byear{2016})
\end{bchapter}
\endbibitem

%%% 34
\bibitem[\protect\citeauthoryear{{H2020 SPRING Project}}{2022}]{spring-d24}
\begin{botherref}
\oauthor{\bsnm{{H2020 SPRING Project}}}:
Deliverable D2.4: Visual-based localisation in relevant environments.
\href{https://spring-h2020.eu/wp-content/uploads/2023/07/SPRING_D2.4_Visual-based-localisation-in-relevant-environments_VFinal_29-07-2022.pdf}{Link}
(2022)
\end{botherref}
\endbibitem

%%% 35
\bibitem[\protect\citeauthoryear{Knapp and Carter}{1976}]{knapp1976generalized}
\begin{barticle}
\bauthor{\bsnm{Knapp}, \binits{C.}},
\bauthor{\bsnm{Carter}, \binits{G.}}:
\batitle{The generalized correlation method for estimation of time delay}.
\bjtitle{IEEE transactions on acoustics, speech, and signal processing}
\bvolume{24}(\bissue{4}),
\bfpage{320}--\blpage{327}
(\byear{1976})
\end{barticle}
\endbibitem

%%% 36
\bibitem[\protect\citeauthoryear{Zhang et~al.}{2021}]{zhang2020fairmot}
\begin{barticle}
\bauthor{\bsnm{Zhang}, \binits{Y.}},
\bauthor{\bsnm{Wang}, \binits{C.}},
\bauthor{\bsnm{Wang}, \binits{X.}},
\bauthor{\bsnm{Zeng}, \binits{W.}},
\bauthor{\bsnm{Liu}, \binits{W.}}:
\batitle{Fairmot: On the fairness of detection and re-identification in
  multiple object tracking}.
\bjtitle{International Journal of Computer Vision}
\bvolume{129},
\bfpage{3069}--\blpage{3087}
(\byear{2021})
\end{barticle}
\endbibitem

%%% 37
\bibitem[\protect\citeauthoryear{He et~al.}{2016}]{he2016deep}
\begin{bchapter}
\bauthor{\bsnm{He}, \binits{K.}},
\bauthor{\bsnm{Zhang}, \binits{X.}},
\bauthor{\bsnm{Ren}, \binits{S.}},
\bauthor{\bsnm{Sun}, \binits{J.}}:
\bctitle{Deep residual learning for image recognition}.
In: \bbtitle{Proceedings of the IEEE Conference on Computer Vision and Pattern
  Recognition},
pp. \bfpage{770}--\blpage{778}
(\byear{2016})
\end{bchapter}
\endbibitem

%%% 38
\bibitem[\protect\citeauthoryear{{H2020 SPRING Project}}{2022}]{spring-d32}
\begin{botherref}
\oauthor{\bsnm{{H2020 SPRING Project}}}:
Deliverable D3.2: Audio-visual speaker tracking in relevant environments.
\href{https://spring-h2020.eu/wp-content/uploads/2023/07/SPRING_D3.2_Audio_Visual_Speaker_Detection_and_Tracking_D3.2_VFinal_29.07.2022.pdf}{Link}
(2022)
\end{botherref}
\endbibitem

%%% 39
\bibitem[\protect\citeauthoryear{Alameda-Pineda
  et~al.}{2011}]{alameda2011finding}
\begin{bchapter}
\bauthor{\bsnm{Alameda-Pineda}, \binits{X.}},
\bauthor{\bsnm{Khalidov}, \binits{V.}},
\bauthor{\bsnm{Horaud}, \binits{R.}},
\bauthor{\bsnm{Forbes}, \binits{F.}}:
\bctitle{Finding audio-visual events in informal social gatherings}.
In: \bbtitle{Proceedings of the 13th International Conference on Multimodal
  Interfaces},
pp. \bfpage{247}--\blpage{254}
(\byear{2011})
\end{bchapter}
\endbibitem

%%% 40
\bibitem[\protect\citeauthoryear{Ban et~al.}{2017}]{ban2017exploiting}
\begin{bchapter}
\bauthor{\bsnm{Ban}, \binits{Y.}},
\bauthor{\bsnm{Girin}, \binits{L.}},
\bauthor{\bsnm{Alameda-Pineda}, \binits{X.}},
\bauthor{\bsnm{Horaud}, \binits{R.}}:
\bctitle{Exploiting the complementarity of audio and visual data in
  multi-speaker tracking}.
In: \bbtitle{Proceedings of the IEEE International Conference on Computer
  Vision Workshops},
pp. \bfpage{446}--\blpage{454}
(\byear{2017})
\end{bchapter}
\endbibitem

%%% 41
\bibitem[\protect\citeauthoryear{{Cao} et~al.}{2019}]{cao2019openpose}
\begin{botherref}
\oauthor{\bsnm{{Cao}}, \binits{Z.}},
\oauthor{\bsnm{{Hidalgo Martinez}}, \binits{G.}},
\oauthor{\bsnm{{Simon}}, \binits{T.}},
\oauthor{\bsnm{{Wei}}, \binits{S.}},
\oauthor{\bsnm{{Sheikh}}, \binits{Y.A.}}:
Openpose: Realtime multi-person 2d pose estimation using part affinity fields.
IEEE Transactions on Pattern Analysis and Machine Intelligence
(2019)
\end{botherref}
\endbibitem

%%% 42
\bibitem[\protect\citeauthoryear{Setti et~al.}{2015}]{setti2015f}
\begin{barticle}
\bauthor{\bsnm{Setti}, \binits{F.}},
\bauthor{\bsnm{Russell}, \binits{C.}},
\bauthor{\bsnm{Bassetti}, \binits{C.}},
\bauthor{\bsnm{Cristani}, \binits{M.}}:
\batitle{F-formation detection: Individuating free-standing conversational
  groups in images}.
\bjtitle{PloS one}
\bvolume{10}(\bissue{5}),
\bfpage{0123783}
(\byear{2015})
\end{barticle}
\endbibitem

%%% 43
\bibitem[\protect\citeauthoryear{Chazan et~al.}{2021}]{Chazan2021MOE}
\begin{bchapter}
\bauthor{\bsnm{Chazan}, \binits{S.E.}},
\bauthor{\bsnm{Goldberger}, \binits{J.}},
\bauthor{\bsnm{Gannot}, \binits{S.}}:
\bctitle{Speech enhancement with mixture of deep experts with clean clustering
  pre-training}.
In: \bbtitle{IEEE International Conference on Audio and Acoustic Signal
  Processing (ICASSP)},
\bconflocation{Toronto, Ontario, Canada}
(\byear{2021})
\end{bchapter}
\endbibitem

%%% 44
\bibitem[\protect\citeauthoryear{Opochinsky et~al.}{2023}]{Opochinsky23single}
\begin{botherref}
\oauthor{\bsnm{Opochinsky}, \binits{R.}},
\oauthor{\bsnm{Moradi1}, \binits{M.}},
\oauthor{\bsnm{Gannot}, \binits{S.}}:
Single-microphone speaker separation and voice activity detection in noisy and
  reverberant environments.
Open Journal on Signal Processing
(2023).
Submitted for publication
\end{botherref}
\endbibitem

%%% 45
\bibitem[\protect\citeauthoryear{Eisenberg et~al.}{2023}]{Eisenberg2023}
\begin{bchapter}
\bauthor{\bsnm{Eisenberg}, \binits{A.}},
\bauthor{\bsnm{Gannot}, \binits{S.}},
\bauthor{\bsnm{Chazan}, \binits{S.E.}}:
\bctitle{A two-stage speaker extraction algorithm under adverse acoustic
  conditions using a single-microphone}.
In: \bbtitle{31st European Signal Processing Conference (EUSIPCO)},
\bconflocation{Helsinki, Finland}
(\byear{2023})
\end{bchapter}
\endbibitem

%%% 46
\bibitem[\protect\citeauthoryear{Dawalatabad
  et~al.}{2021}]{dawalatabad2021ecapa}
\begin{botherref}
\oauthor{\bsnm{Dawalatabad}, \binits{N.}},
\oauthor{\bsnm{Ravanelli}, \binits{M.}},
\oauthor{\bsnm{Grondin}, \binits{F.}},
\oauthor{\bsnm{Thienpondt}, \binits{J.}},
\oauthor{\bsnm{Desplanques}, \binits{B.}},
\oauthor{\bsnm{Na}, \binits{H.}}:
Ecapa-tdnn embeddings for speaker diarization.
arXiv preprint arXiv:2104.01466
(2021)
\end{botherref}
\endbibitem

%%% 47
\bibitem[\protect\citeauthoryear{{H2020 SPRING Project}}{2022}]{spring-d52}
\begin{botherref}
\oauthor{\bsnm{{H2020 SPRING Project}}}:
Deliverable D5.2: Multi-Party ASR and Conversational System in Realistic
  Environments.
\href{https://spring-h2020.eu/wp-content/uploads/2023/07/SPRING_D5.2_Multi_party_ASR_and_conversational_system_VFinal_28-06-2022.pdf}{Link}
(2022)
\end{botherref}
\endbibitem

%%% 48
\bibitem[\protect\citeauthoryear{Chong et~al.}{2020}]{chong2020detecting}
\begin{bchapter}
\bauthor{\bsnm{Chong}, \binits{E.}},
\bauthor{\bsnm{Wang}, \binits{Y.}},
\bauthor{\bsnm{Ruiz}, \binits{N.}},
\bauthor{\bsnm{Rehg}, \binits{J.M.}}:
\bctitle{Detecting attended visual targets in video}.
In: \bbtitle{Proceedings of the IEEE/CVF Conference on Computer Vision and
  Pattern Recognition},
pp. \bfpage{5396}--\blpage{5406}
(\byear{2020})
\end{bchapter}
\endbibitem

%%% 49
\bibitem[\protect\citeauthoryear{Fang et~al.}{2021}]{fang2021dual}
\begin{bchapter}
\bauthor{\bsnm{Fang}, \binits{Y.}},
\bauthor{\bsnm{Tang}, \binits{J.}},
\bauthor{\bsnm{Shen}, \binits{W.}},
\bauthor{\bsnm{Shen}, \binits{W.}},
\bauthor{\bsnm{Gu}, \binits{X.}},
\bauthor{\bsnm{Song}, \binits{L.}},
\bauthor{\bsnm{Zhai}, \binits{G.}}:
\bctitle{Dual attention guided gaze target detection in the wild}.
In: \bbtitle{Proceedings of the IEEE/CVF Conference on Computer Vision and
  Pattern Recognition},
pp. \bfpage{11390}--\blpage{11399}
(\byear{2021})
\end{bchapter}
\endbibitem

%%% 50
\bibitem[\protect\citeauthoryear{Recasens et~al.}{2015}]{Recasens2015}
\begin{bchapter}
\bauthor{\bsnm{Recasens}, \binits{A.}},
\bauthor{\bsnm{Khosla}, \binits{A.}},
\bauthor{\bsnm{Vondrick}, \binits{C.}},
\bauthor{\bsnm{Torralba}, \binits{A.}}:
\bctitle{Where are they looking?}
In: \bbtitle{Advances in Neural Information Processing Systems},
vol. \bseriesno{28}.
\bpublisher{Curran Associates, Inc.}, \blocation{???}
(\byear{2015})
\end{bchapter}
\endbibitem

%%% 51
\bibitem[\protect\citeauthoryear{Tonini et~al.}{2022}]{tonini2022multimodal}
\begin{bchapter}
\bauthor{\bsnm{Tonini}, \binits{F.}},
\bauthor{\bsnm{Beyan}, \binits{C.}},
\bauthor{\bsnm{Ricci}, \binits{E.}}:
\bctitle{Multimodal across domains gaze target detection}.
In: \bbtitle{Proceedings of the 2022 International Conference on Multimodal
  Interaction},
pp. \bfpage{420}--\blpage{431}
(\byear{2022})
\end{bchapter}
\endbibitem

%%% 52
\bibitem[\protect\citeauthoryear{Zhang et~al.}{2016}]{zhang2016joint}
\begin{barticle}
\bauthor{\bsnm{Zhang}, \binits{K.}},
\bauthor{\bsnm{Zhang}, \binits{Z.}},
\bauthor{\bsnm{Li}, \binits{Z.}},
\bauthor{\bsnm{Qiao}, \binits{Y.}}:
\batitle{Joint face detection and alignment using multitask cascaded
  convolutional networks}.
\bjtitle{IEEE signal processing letters}
\bvolume{23}(\bissue{10}),
\bfpage{1499}--\blpage{1503}
(\byear{2016})
\end{barticle}
\endbibitem

%%% 53
\bibitem[\protect\citeauthoryear{Ranftl et~al.}{2020}]{ranftl2020towards}
\begin{botherref}
\oauthor{\bsnm{Ranftl}, \binits{R.}},
\oauthor{\bsnm{Lasinger}, \binits{K.}},
\oauthor{\bsnm{Hafner}, \binits{D.}},
\oauthor{\bsnm{Schindler}, \binits{K.}},
\oauthor{\bsnm{Koltun}, \binits{V.}}:
Towards robust monocular depth estimation: Mixing datasets for zero-shot
  cross-dataset transfer.
IEEE transactions on pattern analysis and machine intelligence
(2020)
\end{botherref}
\endbibitem

%%% 54
\bibitem[\protect\citeauthoryear{Salam and Chetouani}{2015}]{Salam2015}
\begin{bchapter}
\bauthor{\bsnm{Salam}, \binits{H.}},
\bauthor{\bsnm{Chetouani}, \binits{M.}}:
\bctitle{Engagement detection based on mutli-party cues for human robot
  interaction}.
In: \bbtitle{2015 International Conference on Affective Computing and
  Intelligent Interaction (ACII)},
pp. \bfpage{341}--\blpage{347}
(\byear{2015})
\end{bchapter}
\endbibitem

%%% 55
\bibitem[\protect\citeauthoryear{Anzalone
  et~al.}{2015}]{anzalone2015evaluating}
\begin{barticle}
\bauthor{\bsnm{Anzalone}, \binits{S.M.}},
\bauthor{\bsnm{Boucenna}, \binits{S.}},
\bauthor{\bsnm{Ivaldi}, \binits{S.}},
\bauthor{\bsnm{Chetouani}, \binits{M.}}:
\batitle{Evaluating the engagement with social robots}.
\bjtitle{International Journal of Social Robotics}
\bvolume{7},
\bfpage{465}--\blpage{478}
(\byear{2015})
\end{barticle}
\endbibitem

%%% 56
\bibitem[\protect\citeauthoryear{Beyan et~al.}{2017}]{beyan2017prediction}
\begin{barticle}
\bauthor{\bsnm{Beyan}, \binits{C.}},
\bauthor{\bsnm{Capozzi}, \binits{F.}},
\bauthor{\bsnm{Becchio}, \binits{C.}},
\bauthor{\bsnm{Murino}, \binits{V.}}:
\batitle{Prediction of the leadership style of an emergent leader using audio
  and visual nonverbal features}.
\bjtitle{IEEE Transactions on Multimedia}
\bvolume{20}(\bissue{2}),
\bfpage{441}--\blpage{456}
(\byear{2017})
\end{barticle}
\endbibitem

%%% 57
\bibitem[\protect\citeauthoryear{D’inc{\`a} et~al.}{2023}]{d2023unleashing}
\begin{botherref}
\oauthor{\bsnm{D’inc{\`a}}, \binits{M.}},
\oauthor{\bsnm{Beyan}, \binits{C.}},
\oauthor{\bsnm{Niewiadomski}, \binits{R.}},
\oauthor{\bsnm{Barattin}, \binits{S.}},
\oauthor{\bsnm{Sebe}, \binits{N.}}:
Unleashing the transferability power of unsupervised pre-training for emotion
  recognition in masked and unmasked facial images.
IEEE Access
(2023)
\end{botherref}
\endbibitem

%%% 58
\bibitem[\protect\citeauthoryear{Sherman et~al.}{2023}]{Sherman2022}
\begin{bchapter}
\bauthor{\bsnm{Sherman}, \binits{D.}},
\bauthor{\bsnm{Hazan}, \binits{G.}},
\bauthor{\bsnm{Gannot}, \binits{S.}}:
\bctitle{Study of speech emotion recognition using {BLSTM} with attention}.
In: \bbtitle{31st European Signal Processing Conference (EUSIPCO)},
\bconflocation{Helsinki, Finland}
(\byear{2023})
\end{bchapter}
\endbibitem

%%% 59
\bibitem[\protect\citeauthoryear{Huang and Narayanan}{2017}]{huang2017deep}
\begin{bchapter}
\bauthor{\bsnm{Huang}, \binits{C.-W.}},
\bauthor{\bsnm{Narayanan}, \binits{S.S.}}:
\bctitle{Deep convolutional recurrent neural network with attention mechanism
  for robust speech emotion recognition}.
In: \bbtitle{IEEE International Conference on Multimedia and Expo (ICME)},
pp. \bfpage{583}--\blpage{588}
(\byear{2017})
\end{bchapter}
\endbibitem

%%% 60
\bibitem[\protect\citeauthoryear{Bahdanau et~al.}{2014}]{bahdanau2014neural}
\begin{botherref}
\oauthor{\bsnm{Bahdanau}, \binits{D.}},
\oauthor{\bsnm{Cho}, \binits{K.}},
\oauthor{\bsnm{Bengio}, \binits{Y.}}:
Neural machine translation by jointly learning to align and translate.
arXiv preprint arXiv:1409.0473
(2014)
\end{botherref}
\endbibitem

%%% 61
\bibitem[\protect\citeauthoryear{Kuhn}{1955}]{kuhn1955hungarian}
\begin{barticle}
\bauthor{\bsnm{Kuhn}, \binits{H.W.}}:
\batitle{The hungarian method for the assignment problem}.
\bjtitle{Naval research logistics quarterly}
\bvolume{2}(\bissue{1-2}),
\bfpage{83}--\blpage{97}
(\byear{1955})
\end{barticle}
\endbibitem

%%% 62
\bibitem[\protect\citeauthoryear{Lemaignan and
  Ferrini}{2024}]{lemaignan2024mrpotato}
\begin{bchapter}
\bauthor{\bsnm{Lemaignan}, \binits{S.}},
\bauthor{\bsnm{Ferrini}, \binits{L.}}:
\bctitle{Probabilistic fusion of persons' body features: the mr. potato
  algorithm}.
In: \bbtitle{Proceedings of the 2024 ACM/IEEE International Conference on
  Human-Robot Interaction}
(\byear{2024})
\end{bchapter}
\endbibitem

%%% 63
\bibitem[\protect\citeauthoryear{Siek et~al.}{2001}]{siek2001boost}
\begin{bbook}
\bauthor{\bsnm{Siek}, \binits{J.G.}},
\bauthor{\bsnm{Lee}, \binits{L.-Q.}},
\bauthor{\bsnm{Lumsdaine}, \binits{A.}}:
\bbtitle{The Boost Graph Library: User Guide and Reference Manual}.
\bpublisher{Pearson Education}, \blocation{???}
(\byear{2001})
\end{bbook}
\endbibitem

%%% 64
\bibitem[\protect\citeauthoryear{Kruskal}{1956}]{kruskal1956shortest}
\begin{barticle}
\bauthor{\bsnm{Kruskal}, \binits{J.B.}}:
\batitle{On the shortest spanning subtree of a graph and the traveling salesman
  problem}.
\bjtitle{Proceedings of the American Mathematical society}
\bvolume{7}(\bissue{1}),
\bfpage{48}--\blpage{50}
(\byear{1956})
\end{barticle}
\endbibitem

%%% 65
\bibitem[\protect\citeauthoryear{Dijkstra}{2022}]{dijkstra2022note}
\begin{bchapter}
\bauthor{\bsnm{Dijkstra}, \binits{E.W.}}:
\bctitle{A note on two problems in connexion with graphs}.
In: \bbtitle{Edsger Wybe Dijkstra: His Life, Work, and Legacy},
pp. \bfpage{287}--\blpage{290}
(\byear{2022})
\end{bchapter}
\endbibitem

%%% 66
\bibitem[\protect\citeauthoryear{Addlesee
  et~al.}{2023}]{addlesee2023multiparty}
\begin{bchapter}
\bauthor{\bsnm{Addlesee}, \binits{A.}},
\bauthor{\bsnm{Siei{\'n}ska}, \binits{W.}},
\bauthor{\bsnm{Gunson}, \binits{N.}},
\bauthor{\bsnm{Garcia}, \binits{D.H.}},
\bauthor{\bsnm{Dondrup}, \binits{C.}},
\bauthor{\bsnm{Lemon}, \binits{O.}}:
\bctitle{Multi-party goal tracking with llms: Comparing pre-training,
  fine-tuning, and prompt engineering}.
In: \bbtitle{Proceedings of the 24th Annual Meeting of the Special Interest
  Group on Discourse and Dialogue}
(\byear{2023})
\end{bchapter}
\endbibitem

%%% 67
\bibitem[\protect\citeauthoryear{Schauer et~al.}{2023}]{schauer2023detecting}
\begin{bchapter}
\bauthor{\bsnm{Schauer}, \binits{L.}},
\bauthor{\bsnm{Sweeny}, \binits{J.}},
\bauthor{\bsnm{Lyttle}, \binits{C.}},
\bauthor{\bsnm{Said}, \binits{Z.}},
\bauthor{\bsnm{Szeles}, \binits{A.}},
\bauthor{\bsnm{Clark}, \binits{C.}},
\bauthor{\bsnm{McAskill}, \binits{K.}},
\bauthor{\bsnm{Wickham}, \binits{X.}},
\bauthor{\bsnm{Byars}, \binits{T.}},
\bauthor{\bsnm{Garcia}, \binits{D.H.}},
\bauthor{\bsnm{Gunson}, \binits{N.}},
\bauthor{\bsnm{Addlesee}, \binits{A.}},
\bauthor{\bsnm{Lemon}, \binits{O.}}:
\bctitle{Detecting agreement in multi-party conversational ai}.
In: \bbtitle{Proceedings of the Workshop on Advancing GROup UNderstanding and
  Robots aDaptive Behaviour (GROUND)}
(\byear{2023})
\end{bchapter}
\endbibitem

%%% 68
\bibitem[\protect\citeauthoryear{Porcheron et~al.}{2018}]{porcheron2018voice}
\begin{bchapter}
\bauthor{\bsnm{Porcheron}, \binits{M.}},
\bauthor{\bsnm{Fischer}, \binits{J.E.}},
\bauthor{\bsnm{Reeves}, \binits{S.}},
\bauthor{\bsnm{Sharples}, \binits{S.}}:
\bctitle{Voice interfaces in everyday life}.
In: \bbtitle{Proceedings of the 2018 CHI Conference on Human Factors in
  Computing Systems},
pp. \bfpage{1}--\blpage{12}
(\byear{2018})
\end{bchapter}
\endbibitem

%%% 69
\bibitem[\protect\citeauthoryear{Traum}{2004}]{traum2004issues}
\begin{bchapter}
\bauthor{\bsnm{Traum}, \binits{D.}}:
\bctitle{Issues in multiparty dialogues}.
In: \bbtitle{Advances in Agent Communication: International Workshop on Agent
  Communication Languages, ACL 2003, Melbourne, Australia, July 14, 2003.
  Revised and Invited Papers},
pp. \bfpage{201}--\blpage{211}
(\byear{2004}).
\bcomment{Springer}
\end{bchapter}
\endbibitem

%%% 70
\bibitem[\protect\citeauthoryear{Gu et~al.}{2022}]{gu2022says}
\begin{bchapter}
\bauthor{\bsnm{Gu}, \binits{J.-C.}},
\bauthor{\bsnm{Tao}, \binits{C.}},
\bauthor{\bsnm{Ling}, \binits{Z.-H.}}:
\bctitle{{WHO Says WHAT to WHOM: A Survey of Multi-Party Conversations}}.
In: \bbtitle{Proceedings of the Thirty-First International Joint Conference on
  Artificial Intelligence (IJCAI-22)}
(\byear{2022})
\end{bchapter}
\endbibitem

%%% 71
\bibitem[\protect\citeauthoryear{Eshghi and
  Healey}{2016}]{eshghi2016collective}
\begin{barticle}
\bauthor{\bsnm{Eshghi}, \binits{A.}},
\bauthor{\bsnm{Healey}, \binits{P.G.}}:
\batitle{Collective contexts in conversation: Grounding by proxy}.
\bjtitle{Cognitive science}
\bvolume{40}(\bissue{2}),
\bfpage{299}--\blpage{324}
(\byear{2016})
\end{barticle}
\endbibitem

%%% 72
\bibitem[\protect\citeauthoryear{Addlesee et~al.}{2024}]{addlesee2024a}
\begin{bchapter}
\bauthor{\bsnm{Addlesee}, \binits{A.}},
\bauthor{\bsnm{Cherakara}, \binits{N.}},
\bauthor{\bsnm{Nelson}, \binits{N.}},
\bauthor{\bsnm{Garc\'ia}, \binits{D.H.}},
\bauthor{\bsnm{Gunson}, \binits{N.}},
\bauthor{\bsnm{Siei{\'n}ska}, \binits{W.}},
\bauthor{\bsnm{Romeo}, \binits{M.}},
\bauthor{\bsnm{Dondrup}, \binits{C.}},
\bauthor{\bsnm{Lemon}, \binits{O.}}:
\bctitle{A multi-party conversational social robot using llms}.
In: \bbtitle{Companion of the 2024 ACM/IEEE International Conference on
  Human-Robot Interaction (HRI)}
(\byear{2024})
\end{bchapter}
\endbibitem

%%% 73
\bibitem[\protect\citeauthoryear{Papaioannou
  et~al.}{2017}]{papaioannou2017alana}
\begin{botherref}
\oauthor{\bsnm{Papaioannou}, \binits{I.}},
\oauthor{\bsnm{Curry}, \binits{A.C.}},
\oauthor{\bsnm{Part}, \binits{J.L.}},
\oauthor{\bsnm{Shalyminov}, \binits{I.}},
\oauthor{\bsnm{Xu}, \binits{X.}},
\oauthor{\bsnm{Yu}, \binits{Y.}},
\oauthor{\bsnm{Du{\v{s}}ek}, \binits{O.}},
\oauthor{\bsnm{Rieser}, \binits{V.}},
\oauthor{\bsnm{Lemon}, \binits{O.}}:
Alana: Social dialogue using an ensemble model and a ranker trained on user
  feedback.
Alexa Prize Proceedings
(2017)
\end{botherref}
\endbibitem

%%% 74
\bibitem[\protect\citeauthoryear{Curry et~al.}{2018}]{curry2018alana}
\begin{botherref}
\oauthor{\bsnm{Curry}, \binits{A.C.}},
\oauthor{\bsnm{Papaioannou}, \binits{I.}},
\oauthor{\bsnm{Suglia}, \binits{A.}},
\oauthor{\bsnm{Agarwal}, \binits{S.}},
\oauthor{\bsnm{Shalyminov}, \binits{I.}},
\oauthor{\bsnm{Xu}, \binits{X.}},
\oauthor{\bsnm{Du{\v{s}}ek}, \binits{O.}},
\oauthor{\bsnm{Eshghi}, \binits{A.}},
\oauthor{\bsnm{Konstas}, \binits{I.}},
\oauthor{\bsnm{Rieser}, \binits{V.}}, et al.:
Alana v2: Entertaining and informative open-domain social dialogue using
  ontologies and entity linking.
Alexa Prize Proceedings
(2018)
\end{botherref}
\endbibitem

%%% 75
\bibitem[\protect\citeauthoryear{Addlesee et~al.}{2023a}]{addlesee2023data}
\begin{bchapter}
\bauthor{\bsnm{Addlesee}, \binits{A.}},
\bauthor{\bsnm{Siei{\'n}ska}, \binits{W.}},
\bauthor{\bsnm{Gunson}, \binits{N.}},
\bauthor{\bsnm{Garcia}, \binits{D.H.}},
\bauthor{\bsnm{Dondrup}, \binits{C.}},
\bauthor{\bsnm{Lemon}, \binits{O.}}:
\bctitle{Data collection for multi-party task-based dialogue in social
  robotics}.
In: \bbtitle{Proceedings of the 13th International Workshop on Spoken Dialogue
  Systems Technology (IWSDS)}
(\byear{2023})
\end{bchapter}
\endbibitem

%%% 76
\bibitem[\protect\citeauthoryear{Addlesee
  et~al.}{2023b}]{addlesee2023detecting}
\begin{bchapter}
\bauthor{\bsnm{Addlesee}, \binits{A.}},
\bauthor{\bsnm{Denley}, \binits{D.}},
\bauthor{\bsnm{Edmondson}, \binits{A.}},
\bauthor{\bsnm{Gunson}, \binits{N.}},
\bauthor{\bsnm{Garcia}, \binits{D.H.}},
\bauthor{\bsnm{Kha}, \binits{A.}},
\bauthor{\bsnm{Lemon}, \binits{O.}},
\bauthor{\bsnm{Ndubuisi}, \binits{J.}},
\bauthor{\bsnm{O’Reilly}, \binits{N.}},
\bauthor{\bsnm{Perochaud}, \binits{L.}},
\bauthor{\bsnm{Valeri}, \binits{R.}},
\bauthor{\bsnm{Worika}, \binits{M.}}:
\bctitle{Detecting agreement in multi-party dialogue: evaluating speaker
  diarisation versus a procedural baseline to enhance user engagement}.
In: \bbtitle{Proceedings of the Workshop on Advancing GROup UNderstanding and
  Robots aDaptive Behaviour (GROUND)}
(\byear{2023})
\end{bchapter}
\endbibitem

%%% 77
\bibitem[\protect\citeauthoryear{Dondrup et~al.}{2019}]{Dondrup2019}
\begin{botherref}
\oauthor{\bsnm{Dondrup}, \binits{C.}},
\oauthor{\bsnm{Papaioannou}, \binits{I.}},
\oauthor{\bsnm{Lemon}, \binits{O.}}:
{Petri Net Machines for Human-Agent Interaction}
(2019)
\end{botherref}
\endbibitem

%%% 78
\bibitem[\protect\citeauthoryear{Kruse et~al.}{2013}]{kruse2013human}
\begin{barticle}
\bauthor{\bsnm{Kruse}, \binits{T.}},
\bauthor{\bsnm{Pandey}, \binits{A.K.}},
\bauthor{\bsnm{Alami}, \binits{R.}},
\bauthor{\bsnm{Kirsch}, \binits{A.}}:
\batitle{Human-aware robot navigation: A survey}.
\bjtitle{Robotics and Autonomous Systems}
\bvolume{61}(\bissue{12}),
\bfpage{1726}--\blpage{1743}
(\byear{2013})
\end{barticle}
\endbibitem

%%% 79
\bibitem[\protect\citeauthoryear{Mavrogiannis
  et~al.}{2023}]{mavrogiannis2023core}
\begin{barticle}
\bauthor{\bsnm{Mavrogiannis}, \binits{C.}},
\bauthor{\bsnm{Baldini}, \binits{F.}},
\bauthor{\bsnm{Wang}, \binits{A.}},
\bauthor{\bsnm{Zhao}, \binits{D.}},
\bauthor{\bsnm{Trautman}, \binits{P.}},
\bauthor{\bsnm{Steinfeld}, \binits{A.}},
\bauthor{\bsnm{Oh}, \binits{J.}}:
\batitle{Core challenges of social robot navigation: A survey}.
\bjtitle{ACM Transactions on Human-Robot Interaction}
\bvolume{12}(\bissue{3}),
\bfpage{1}--\blpage{39}
(\byear{2023})
\end{barticle}
\endbibitem

%%% 80
\bibitem[\protect\citeauthoryear{Singamaneni
  et~al.}{2023}]{singamaneni2023survey}
\begin{botherref}
\oauthor{\bsnm{Singamaneni}, \binits{P.T.}},
\oauthor{\bsnm{Bachiller-Burgos}, \binits{P.}},
\oauthor{\bsnm{Manso}, \binits{L.J.}},
\oauthor{\bsnm{Garrell}, \binits{A.}},
\oauthor{\bsnm{Sanfeliu}, \binits{A.}},
\oauthor{\bsnm{Spalanzani}, \binits{A.}},
\oauthor{\bsnm{Alami}, \binits{R.}}, et al.:
A survey on socially aware robot navigation: Taxonomy and future challenges.
arXiv preprint arXiv:2311.06922
(2023)
\end{botherref}
\endbibitem

%%% 81
\bibitem[\protect\citeauthoryear{Camacho and Bordons}{2007}]{camacho2007model}
\begin{bbook}
\bauthor{\bsnm{Camacho}, \binits{E.F.}},
\bauthor{\bsnm{Bordons}, \binits{C.}}:
\bbtitle{Model Predictive Control}.
\bpublisher{Springer},
\blocation{London}
(\byear{2007})
\end{bbook}
\endbibitem

%%% 82
\bibitem[\protect\citeauthoryear{Truong and Ngo}{2016}]{truong2016dynamic}
\begin{barticle}
\bauthor{\bsnm{Truong}, \binits{X.-T.}},
\bauthor{\bsnm{Ngo}, \binits{T.-D.}}:
\batitle{Dynamic social zone based mobile robot navigation for human
  comfortable safety in social environments}.
\bjtitle{International Journal of Social Robotics}
\bvolume{8}(\bissue{5}),
\bfpage{663}--\blpage{684}
(\byear{2016})
\end{barticle}
\endbibitem

%%% 83
\bibitem[\protect\citeauthoryear{{H2020 SPRING Project}}{2022}]{spring-d63}
\begin{botherref}
\oauthor{\bsnm{{H2020 SPRING Project}}}:
Deliverable D6.3: Robot non-verbal behaviour system in realistic environments.
\href{https://spring-h2020.eu/wp-content/uploads/2022/02/SPRING_D6.3_Robot_non-verbal_behaviour_system_in_realistic_environments_VFinal_25.01.2022.pdf}{Link}
(2022)
\end{botherref}
\endbibitem

%%% 84
\bibitem[\protect\citeauthoryear{Truong and Ngo}{2017}]{truong2017approach}
\begin{barticle}
\bauthor{\bsnm{Truong}, \binits{X.-T.}},
\bauthor{\bsnm{Ngo}, \binits{T.-D.}}:
\batitle{“to approach humans?”: A unified framework for approaching pose
  prediction and socially aware robot navigation}.
\bjtitle{IEEE Transactions on Cognitive and Developmental Systems}
\bvolume{10}(\bissue{3}),
\bfpage{557}--\blpage{572}
(\byear{2017})
\end{barticle}
\endbibitem

%%% 85
\bibitem[\protect\citeauthoryear{Chen et~al.}{2020}]{chen2020robot}
\begin{bchapter}
\bauthor{\bsnm{Chen}, \binits{G.}},
\bauthor{\bsnm{Pan}, \binits{L.}},
\bauthor{\bsnm{Xu}, \binits{P.}},
\bauthor{\bsnm{Wang}, \binits{Z.}},
\bauthor{\bsnm{Wu}, \binits{P.}},
\bauthor{\bsnm{Ji}, \binits{J.}},
\bauthor{\bsnm{Chen}, \binits{X.}}, \betal:
\bctitle{Robot navigation with map-based deep reinforcement learning}.
In: \bbtitle{2020 IEEE International Conference on Networking, Sensing and
  Control (ICNSC)},
pp. \bfpage{1}--\blpage{6}
(\byear{2020}).
\bcomment{IEEE}
\end{bchapter}
\endbibitem

%%% 86
\bibitem[\protect\citeauthoryear{{H2020 SPRING Project}}{2023}]{spring-d53}
\begin{botherref}
\oauthor{\bsnm{{H2020 SPRING Project}}}:
Deliverable D5.3: High-Level task planner in relevant environments.
\href{https://spring-h2020.eu/wp-content/uploads/2023/07/SPRING_D5_3_High_Level_task_planner_in_relevant_environments_VFinal_28.03.2023-1.pdf}{Link}
(2023)
\end{botherref}
\endbibitem

%%% 87
\bibitem[\protect\citeauthoryear{Folstein et~al.}{1975}]{folstein1975mini}
\begin{barticle}
\bauthor{\bsnm{Folstein}, \binits{M.F.}},
\bauthor{\bsnm{Folstein}, \binits{S.E.}},
\bauthor{\bsnm{McHugh}, \binits{P.R.}}:
\batitle{“mini-mental state”: a practical method for grading the cognitive
  state of patients for the clinician}.
\bjtitle{Journal of psychiatric research}
\bvolume{12}(\bissue{3}),
\bfpage{189}--\blpage{198}
(\byear{1975})
\end{barticle}
\endbibitem

%%% 88
\bibitem[\protect\citeauthoryear{{H2020 SPRING Project}}{2020}]{spring-d12}
\begin{botherref}
\oauthor{\bsnm{{H2020 SPRING Project}}}:
Deliverable D1.2: Privacy and Ethics guidelines for experimental validation and
  data collection.
\href{https://spring-h2020.eu/wp-content/uploads/2023/07/SPRING_D1.2_Privacy-and-Ethics-Guidelines_VFinal_Amended_29.06.2022.pdf}{Link}
(2020)
\end{botherref}
\endbibitem

%%% 89
\bibitem[\protect\citeauthoryear{Micoulaud-Franchi
  et~al.}{2016}]{micoulaud2016validation}
\begin{barticle}
\bauthor{\bsnm{Micoulaud-Franchi}, \binits{J.-A.}},
\bauthor{\bsnm{Sauteraud}, \binits{A.}},
\bauthor{\bsnm{Olive}, \binits{J.}},
\bauthor{\bsnm{Sagaspe}, \binits{P.}},
\bauthor{\bsnm{Bioulac}, \binits{S.}},
\bauthor{\bsnm{Philip}, \binits{P.}}:
\batitle{Validation of the french version of the acceptability e-scale (aes)
  for mental e-health systems}.
\bjtitle{Psychiatry Research}
\bvolume{237},
\bfpage{196}--\blpage{200}
(\byear{2016})
\end{barticle}
\endbibitem

%%% 90
\bibitem[\protect\citeauthoryear{Brooke}{1996}]{brooke1996system}
\begin{botherref}
\oauthor{\bsnm{Brooke}, \binits{J.}}:
System Usability Scale (SUS): A Quick-and-Dirty Method of System Evaluation
  User Information. Usability Evaluation In Industry
(1996)
\end{botherref}
\endbibitem

%%% 91
\bibitem[\protect\citeauthoryear{Saplacan et~al.}{2023}]{saplacan2023health}
\begin{bchapter}
\bauthor{\bsnm{Saplacan}, \binits{D.}},
\bauthor{\bsnm{Schulz}, \binits{T.}},
\bauthor{\bsnm{Torresen}, \binits{J.}},
\bauthor{\bsnm{Pajalic}, \binits{Z.}}:
\bctitle{Health professionals’ views on the use of social robots with
  vulnerable users: A scenario-based qualitative study using story dialogue
  method}.
In: \bbtitle{IEEE International Conference on Robot and Human Interactive
  Communication (RO-MAN)},
pp. \bfpage{421}--\blpage{428}
(\byear{2023})
\end{bchapter}
\endbibitem

%%% 92
\bibitem[\protect\citeauthoryear{Orinel and Constant}{2021}]{orinel2021dossier}
\begin{barticle}
\bauthor{\bsnm{Orinel}, \binits{F.}},
\bauthor{\bsnm{Constant}, \binits{N.}}:
\batitle{Le dossier de soins {\`a} l’{\`e}re du num{\'e}rique}.
\bjtitle{L'Aide-Soignante}
\bvolume{35}(\bissue{231}),
\bfpage{13}--\blpage{14}
(\byear{2021})
\end{barticle}
\endbibitem

%%% 93
\bibitem[\protect\citeauthoryear{Beaugrand}{1988}]{beaugrand1988demarche}
\begin{botherref}
\oauthor{\bsnm{Beaugrand}, \binits{J.P.}}:
D{\'e}marche scientifique et cycle de la recherche.
Fondements et {\'e}tapes de la recherche scientifique en psychologie,
1--35
(1988)
\end{botherref}
\endbibitem

%%% 94
\bibitem[\protect\citeauthoryear{Lindblom
  et~al.}{2024}]{lindblom2024qualitative}
\begin{bchapter}
\bauthor{\bsnm{Lindblom}, \binits{D.S.}},
\bauthor{\bsnm{Otterdijk}, \binits{M.}},
\bauthor{\bsnm{Torresen}, \binits{J.}}:
\bctitle{A qualitative observational video-based study on perceived privacy in
  social robots’ based on robots appearances}.
In: \bbtitle{IEEE International Conference on Advanced Robotics and Its Social
  Impacts (ARSO)},
pp. \bfpage{74}--\blpage{79}
(\byear{2024})
\end{bchapter}
\endbibitem

%%% 95
\bibitem[\protect\citeauthoryear{Saplacan et~al.}{2021}]{saplacan2021ethical}
\begin{bchapter}
\bauthor{\bsnm{Saplacan}, \binits{D.}},
\bauthor{\bsnm{Khaksar}, \binits{W.}},
\bauthor{\bsnm{Torresen}, \binits{J.}}:
\bctitle{On ethical challenges raised by care robots: a review of the existing
  regulatory-, theoretical-, and research gaps}.
In: \bbtitle{IEEE International Conference on Advanced Robotics and Its Social
  Impacts (ARSO)},
pp. \bfpage{219}--\blpage{226}
(\byear{2021})
\end{bchapter}
\endbibitem

%%% 96
\bibitem[\protect\citeauthoryear{Pedersen
  et~al.}{2018}]{pedersen2018developing}
\begin{barticle}
\bauthor{\bsnm{Pedersen}, \binits{I.}},
\bauthor{\bsnm{Reid}, \binits{S.}},
\bauthor{\bsnm{Aspevig}, \binits{K.}}:
\batitle{Developing social robots for aging populations: A literature review of
  recent academic sources}.
\bjtitle{Sociology Compass}
\bvolume{12}(\bissue{6}),
\bfpage{12585}
(\byear{2018})
\end{barticle}
\endbibitem

%%% 97
\bibitem[\protect\citeauthoryear{G{\'o}ngora~Alonso
  et~al.}{2019}]{gongora2019social}
\begin{barticle}
\bauthor{\bsnm{G{\'o}ngora~Alonso}, \binits{S.}},
\bauthor{\bsnm{Hamrioui}, \binits{S.}},
\bauthor{\bsnm{Torre~D{\'\i}ez}, \binits{I.}},
\bauthor{\bsnm{Motta~Cruz}, \binits{E.}},
\bauthor{\bsnm{L{\'o}pez-Coronado}, \binits{M.}},
\bauthor{\bsnm{Franco}, \binits{M.}}:
\batitle{Social robots for people with aging and dementia: a systematic review
  of literature}.
\bjtitle{Telemedicine and e-Health}
\bvolume{25}(\bissue{7}),
\bfpage{533}--\blpage{540}
(\byear{2019})
\end{barticle}
\endbibitem

\end{thebibliography}

\renewcommand{\thesection}{\Alph{section}}
\setcounter{section}{0}
\section{AES and SUS scales}
\label{sec:aes-sus}
\addnote[aes]{1}{The items of the  Acceptability E-Scale (AES) are the following:
\begin{enumerate}
\item To what extent do you find the robot easy to use (use, understand, act)?
\item To what extent do the requests or instructions from the robot seem easy to understand?
\item To what extent did you enjoy using the robot (use, understand, act)?
\item To what extent do you think you will need help to use the robot?
\item Do you think the robot services (features) have been well designed?
\item To what extent are you satisfied with the robot?
\end{enumerate}
The five possible scores for each of the items above are:
\begin{enumerate}
\item Not at all
\item A little
\item Moderately
\item Quite a bit
\item Extremely    
\end{enumerate}}

\addnote[sus]{1}{The items of the System Usability Scale (SUS) are the following:
\begin{enumerate}
\item I would like to use this robot at the hospital as frequently as possible.
\item I find conversations with this robot unnecessarily complex.
\item I think this robot is easy to use (talk, behave, act).
\item I think I would need help to be able to interact with this robot.
\item I found that the various services of this robot were well thought out.
\item I think there are too many inconsistencies with this robot.
\item I imagine that most people would be able to learn how to use this robot very quickly.
\item I found it very difficult to talk and behave naturally with this robot.
\item I felt very confident while using the robot.
\end{enumerate}
The five possible scores for each of the items above are:
\begin{enumerate}
\item Strongly disagree
\item Disagree
\item Neutral
\item Agree
\item Strongly agree
\end{enumerate}
}

\begin{table*}[p!] 
    \centering
    \caption{The complete list of author contributions. $^\dagger$ indicates equal contribution of the first four authors, $^*$ indicates the corresponding author, and the numeric superscripts correspond to the affiliation as shown in the title page.}
    \label{tab:authors}
    \resizebox{0.78\textwidth}{!}{\begin{tabular}{ ll c c c c c c c c c c c c c c }
        \multicolumn{1}{l}{Author}  & 
        \multicolumn{1}{l}{Affiliation} &
        \mcrot{1}{l}{60}{Conceptualization} &
        \mcrot{1}{l}{60}{Data curation} &
        \mcrot{1}{l}{60}{Formal Analysis} &
        \mcrot{1}{l}{60}{Funding acquisition} &
        \mcrot{1}{l}{60}{Investigation} &
        \mcrot{1}{l}{60}{Methodology} &
        \mcrot{1}{l}{60}{Project administration} &
        \mcrot{1}{l}{60}{Resources} &
        \mcrot{1}{l}{60}{Software} &
        \mcrot{1}{l}{60}{Supervision} &
        \mcrot{1}{l}{60}{Validation} &
        \mcrot{1}{l}{60}{Visualization} &
        \mcrot{1}{l}{60}{Writing – original draft} &
        \mcrot{1}{l}{60}{Writing – review \& editing}
        % \mcrot{1}{l}{60}{\twoelementtable{Writing}{~~review \& editing}} & \phantom{p} 
        \\

    \midrule 
    \midrule 

Xavier Alameda-Pineda$^{\dagger*}$ & INRIA$^1$ & x & x & - & x & - & x & x & - & - & x & - & - & x & x\\
Angus Addlesee$^\dagger$ & HWU$^5$                & x & x & x & - & x & x & - & - & x & - & - & x & x & x\\
Daniel Hern\'andez Garc\'ia$^\dagger$ & HWU$^5$   & x & x & - & - & x & x & - & - & x & - & x & x & x & x\\
Chris Reinke$^\dagger$ & INRIA$^1$                & x & - & x & - & - & x & - & - & x & x & - & - & x & x\\
Soraya Arias & INRIA$^1$                    & - & - & - & - & - & - & - & x & - & - & - & - & - & -\\
Federica Arrigoni & UNITN$^4$                & x & - & - & - & - & x & - & - & - & - & - & - & - & -\\
Alex Auternaud & INRIA$^1$                  & - & - & - & - & - & - & - & x & x & - & x & - & - & -\\
Lauriane Blavette & AP-HP$^8$                & - & - & x & - & x & - & - & - & - & - & - & - & x & -\\
Cigdem Beyan & UNITN$^4$                     & x & - & - & - & - & x & - & - & x & - & x & - & x & x\\
Luis Gomez Camara & INRIA$^1$               & - & - & - & - & - & x & - & x & x & - & x & - & - & -\\
Ohad Cohen & BIU$^3$                        & - & - & - & - & x & - & - & - & x & - & - & - & x & -\\
Alessandro Conti & UNITN$^4$                 & - & - & - & - & - & x & - & - & x & - & x & - & - & -\\
Sébastien Dacunha & AP-HP$^8$                & - & - & - & - & - & - & - & - & - & - & x & - & - & x\\
Christian Dondrup & HWU$^5$                 & x & - & - & x & - & x & x & x & x & x & - & - & - & -\\
Yoav Ellinson & BIU$^3$                     & - & - & - & - & x & x & - & - & x & - & - & - & x & -\\
Francesco Ferro & PAL$^7$                   & - & - & - & x & - & - & x & x & - & - & - & - & - & -\\
Sharon Gannot & BIU$^3$                     & x & - & - & x & x & x & x & - & - & x & x & - & - & x\\
Florian Gras & ERM$^6$                      & x & x & - & - & - & - & - & - & x & - & x & - & - & -\\
Nancie Gunson & HWU$^5$                     & x & x & x & - & x & - & - & - & x & - & x & - & - & -\\
Radu Horaud & INRIA$^1$                     & x & - & - & x & - & - & - & - & - & - & - & - & - & -\\
Moreno D'Incà & UNITN$^4$                    & - & - & - & - & - & - & - & - & x & - & x & - & - & -\\
Imad Kimouche & ERM$^6$                     & - & x & - & - & - & - & - & - & x & - & x & - & - & -\\
Séverin Lemaignan & PAL$^7$                 & x & - & - & - & - & x & - & - & x & x & - & - & x & -\\
%                                       & C & D & F & F & I & M & P & R & S & S & V & V & W & W\\
Oliver Lemon & HWU$^5$                      & x & - & - & x & x & x & x & x & - & x & x & - & - & x\\
Cyril Liotard & ERM$^6$                     & x & - & - & x & - & - & - & - & - & x & - & - & - & -\\
Luca Marchionni & PAL$^7$                   & x & - & - & x & - & - & - & x & - & - & - & - & - & -\\
Mordehay Moradi & BIU$^3$                   & - & - & - & - & x & - & - & - & x & - & - & - & x & -\\
Tomas Pajdla & CVUT$^2$                     & x & - & x & x & - & x & x & x & - & x & - & - & - & x\\
Maribel Pino & AP-HP$^8$                     & - & - & - & - & - & x & - & - & - & x & x & - & - & x\\
Michal Polic & CVUT$^2$                     & x & - & x & - & x & x & x & - & x & x & x & x & x & -\\
Matthieu Py & INRIA$^1$                     & - & - & - & - & - & - & x & - & - & - & - & - & - & x\\
Ariel Rado & BIU$^3$                        & - & - & - & - & x & - & - & - & x & - & - & - & x & -\\
Bin Ren & UNITN$^4$                          & - & - & - & - & - & - & - & - & x & - & x & - & - & -\\
Elisa Ricci & UNITN$^4$                      & x & - & - & x & - & - & x & - & - & x & - & - & - & -\\
Anne-Sophie Rigaud & AP-HP$^8$               & - & - & - & - & - & x & - & - & - & x & x & - & - & x\\
Paolo Rota & UNITN$^4$                       & x & - & - & - & - & x & - & - & - & - & - & - & - & -\\
Marta Romeo & HWU$^5$                       & x & - & - & - & - & x & - & - & x & - & x & - & - & -\\
Nicu Sebe & UNITN$^4$                        & - & - & - & x & - & - & x & - & - & x & - & - & - & -\\
Weronika Siei\'nska & HWU$^5$                 & - & x & x & - & - & - & - & - & x & - & - & - & - & -\\
Pinchas Tandeitnik & BIU$^3$                & - & - & - & - & x & x & x & - & - & x & x & - & - & x\\
Francesco Tonini & UNITN$^4$                 & - & - & - & - & - & x & - & - & x & - & x & - & x & x\\
Nicolas Turro & INRIA$^1$                   & - & - & - & - & - & - & - & x & x & - & x & - & - & -\\
Timothée Wintz & INRIA$^1$                  & x & - & - & - & x & x & - & - & x & x & - & - & - & -\\
Yanchao Yu & HWU$^5$                        & x & - & - & - & - & - & - & - & x & - & x & - & - & -\\
    \bottomrule
    \end{tabular}}
\end{table*}

\end{document}